%% file: neurips_2026.tex
\documentclass{article}

\usepackage[final,eandd]{neurips_2026}

\usepackage[utf8]{inputenc}
\usepackage[T1]{fontenc}
\usepackage{hyperref}
\usepackage{url}
\usepackage{booktabs}
\usepackage{amsfonts}
\usepackage{amsmath}
\usepackage{nicefrac}
\usepackage{microtype}
\usepackage[table]{xcolor}
\usepackage{graphicx}
\usepackage{subcaption}
\usepackage{pifont}
\usepackage{fontawesome5}
\newcommand{\cmark}{\ding{51}}
\newcommand{\xmark}{\ding{55}}

\definecolor{website_color}{HTML}{E91E63}
\definecolor{trophygold}{HTML}{D4A017}

\usepackage{tikz}
\usetikzlibrary{arrows.meta,positioning,calc,fit,backgrounds,shapes.geometric}
\usepackage[most]{tcolorbox}
\newtcolorbox{questionbox}[1]{
  enhanced,
  colback=gray!8,
  colframe=gray!40,
  colbacktitle=gray!20,
  coltitle=black,
  fonttitle=\bfseries,
  title={#1},
  boxrule=0.4pt,
  arc=2pt,
  left=8pt, right=8pt, top=4pt, bottom=4pt,
  breakable
}

\newtcolorbox{findingbox}[1]{enhanced, breakable, colback=green!4, colframe=green!45!black, colbacktitle=green!10, coltitle=black, fonttitle=\bfseries\small, title={#1}, boxrule=0.5pt, arc=2pt, left=6pt, right=6pt, top=2pt, bottom=2pt, toptitle=1pt, bottomtitle=1pt, fontupper=\small, before skip=4pt, after skip=4pt}

\graphicspath{{figures/}}

\title{EgoSurgHands: An Egocentric 3D Hand Pose Dataset \& Benchmark for open surgery Training}

\author{%
  Ankit Pal$^{1}$ \quad Guillaume Kugener$^{2}$ \quad Sofia Volpi$^{2}$ \quad Joshua Schwartz$^{2}$ \\
  \textbf{Alex Idarraga$^{2}$ \quad Pranav Rajpurkar$^{1,\dagger}$ \quad Gabriel A Brat$^{1,2,\dagger}$} \\[4pt]
  $^{1}$Department of Biomedical Informatics, Harvard Medical School, Boston, MA, USA \\
  $^{2}$Department of Surgery, Beth Israel Deaconess Medical Center, Boston, MA, USA \\
  $^{\dagger}$Equal senior authorship \\[2pt]
  \texttt{ankit\_pal@fas.harvard.edu} \\
  \texttt{pranav\_rajpurkar@hms.harvard.edu} \quad \texttt{gbrat@bidmc.harvard.edu}
}

\begin{document}

\maketitle
\vspace{-1.5em}
\begin{center}
\hypersetup{hidelinks}%
\raisebox{-0.2em}{\includegraphics[height=1em]{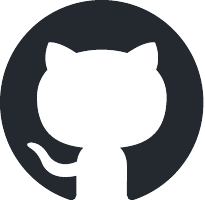}}\hspace{0.4em}\textbf{Code}\hspace{0.6em}\href{https://github.com/monk1337/egoSurgHands}{\textcolor{website_color}{github.com/monk1337/egoSurgHands}}\\[2pt]
\raisebox{-0.05em}{\textcolor{trophygold}{\faTrophy}}\hspace{0.4em}\textbf{Leaderboard \& Project page}\hspace{0.6em}\href{https://egosurghands.github.io}{\textcolor{website_color}{egosurghands.github.io}}\\[2pt]
\raisebox{-0.2em}{\includegraphics[height=1em]{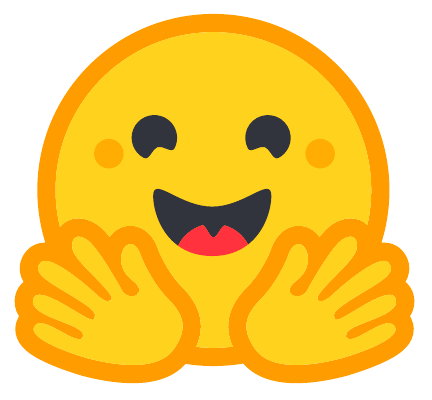}}\hspace{0.4em}\textbf{Dataset}\hspace{0.6em}\href{https://huggingface.co/datasets/surgicalinformatics/EgoSurgHands}{\textcolor{website_color}{huggingface.co/datasets/surgicalinformatics/EgoSurgHands}}
\end{center}

\begin{figure}[h]
  \centering
  \includegraphics[width=0.9\linewidth]{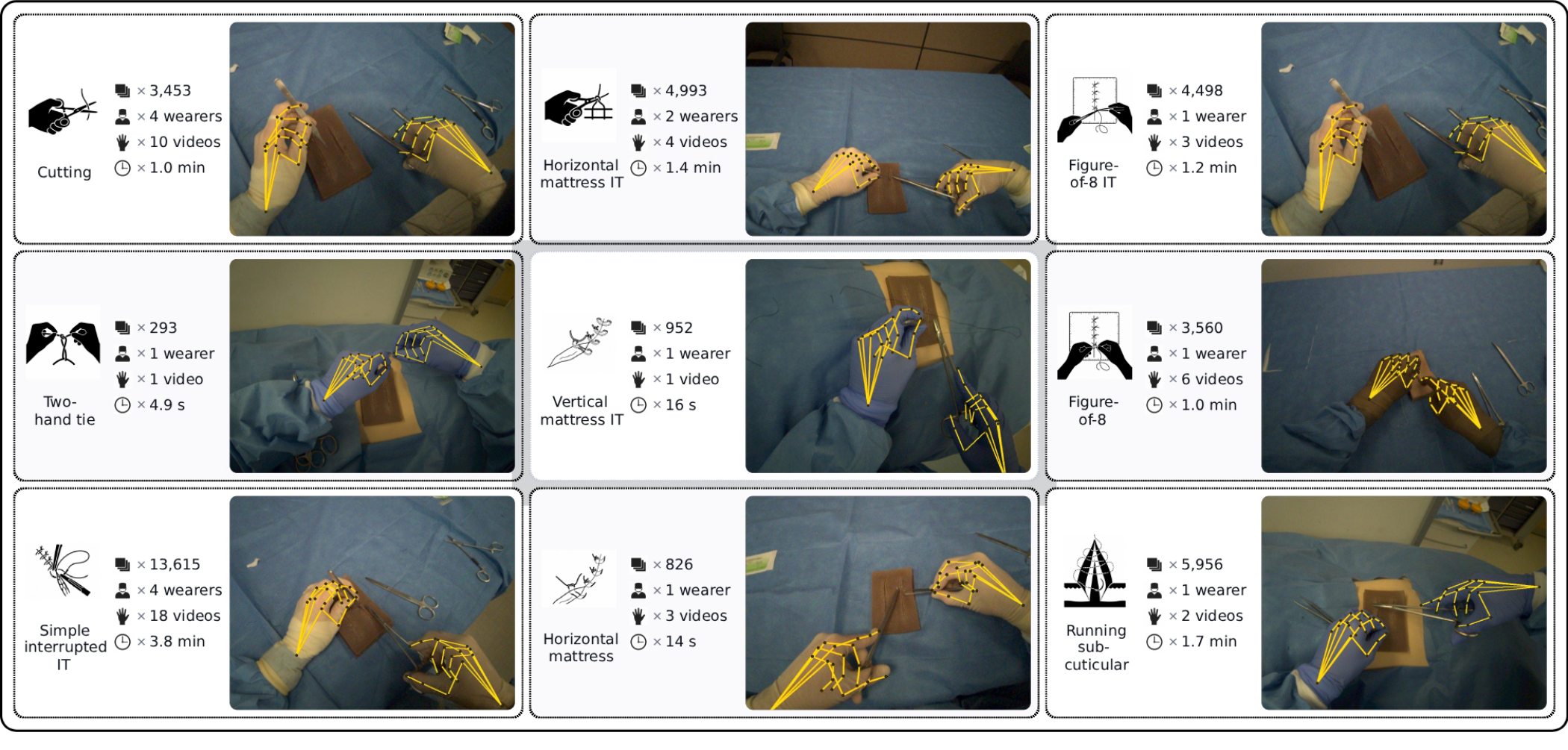}
  \caption{\textbf{\small EgoSurgHands at a glance.} \footnotesize 51{,}385 egocentric hand instances in 32{,}646 frames, with 21-joint 3D labels (9{,}476 human-validated for evaluation) across 8 surgical training task families (13 task variants), 4 wearers (P1-P4), 3 experience tiers (MS-1, PGY-0, PGY-3), and 3 glove colors, captured with egocentric glasses (1408$\times$1408 fisheye RGB at 30 fps).}
  \label{fig:teaser}
\end{figure}

\begin{abstract}
Surgical skill assessment, AR guidance, and robot learning from demonstration all benefit from 3D tracking of the surgeon's hands, yet no surgical dataset offers real egocentric articulated 3D hand pose. We introduce EgoSurgHands, a benchmark for real-captured egocentric hand pose estimation in open surgery training on synthetic pads. Captured with multi-sensor glasses, it contains 41,909 training hand-instance rows across 55 procedure recordings and 9,476 human-validated evaluation rows across 13 held-out recordings, spanning 13 tasks, 3 glove colors, and 4 trainees from first-year medical student to third-year resident. Each frame carries a multi-sensor stack: intrinsics, hand landmarks with per-hand confidence, timestamps, and eye gaze, plus SLAM extrinsics where SLAM succeeded. Evaluation labels are sensor-fused with human-validated pseudo-ground truth: every evaluation frame passes human review with examiner cross-review and surgeon adjudication of flagged cases, and an external multi-view triangulated reference. Alongside the within-corpus, procedure-disjoint split, a leave-one-wearer-out protocol tested unseen trainees. Across eight methods evaluated zero-shot, from MANO-based reconstructors to the two-hand IntagHand and the keypoint-only MediaPipe Hands, egocentric pretraining was associated with stronger transfer on every held-out wearer.  WildHands and JointTransformer had 18\% and 40\% lower PA-MPJPE, respectively, than the non-egocentric HaMeR. JointTransformer was the best performing model with an error of 14.95 mm. A substantial surgical domain gap remains, and the widely deployed MediaPipe Hands detected a hand in only 24.1\% of evaluation rows. As a lightweight diagnostic baseline, a 58K-parameter residual corrector (JRN) on frozen backbones reached 10.96-14.85 mm PA-MPJPE on held-out wearers for the two ego-pretrained backbones. We release the dataset and benchmark code at \url{https://huggingface.co/datasets/surgicalinformatics/EgoSurgHands} and \url{https://github.com/monk1337/egoSurgHands}.

\end{abstract}

\section{Introduction}
\label{sec:intro}

Egocentric 3D hand pose is a key perceptual signal for surgical skill assessment, AR-guided training, and robot learning from demonstration: in suturing, a surgeon's hands encode grip choice, suture tension, needle trajectory, and the sequencing of one-hand and two-hand ties. Large egocentric hand datasets now support robot learning of dexterous manipulation from human video~\cite{hoque2026egodex}, but none covers surgery, and progress there is limited by data. Hand-pose backbones are largely trained on kitchens, factories, sports, or assembly tasks and do not transfer reliably to surgical settings, where colored gloves, surgical drapes, occlusion, and metallic instruments degrade the visual cues that vision-only models depend on.

\textbf{Existing egocentric hand-pose datasets do not cover surgical training.} EgoExo4D~\cite{egoexo4d}, ARCTIC~\cite{arctic}, AssemblyHands~\cite{assemblyhands}, and HOT3D~\cite{hot3d} capture skilled activities, toy assembly, and object manipulation, while surgical datasets are non-egocentric (AIxSuture~\cite{aixsuture}; the multi-view replica-operating-room benchmark of \citet{fischer2026multiview}), synthetic (POV-Surgery~\cite{povsurgery}), or limited to wrist and palm tracking (EgoExOR~\cite{egoexor}). No public benchmark is simultaneously real, egocentric, sensor-grounded, and human-validated for open surgery training tasks.

We introduce \textbf{EgoSurgHands}, the first benchmark for real-captured egocentric hand pose estimation in open surgery training on synthetic pads. Captured with Meta Project Aria glasses, it provides 41{,}909 training hand-instance rows across 55 procedure recordings (PRs) and 9{,}476 human-validated evaluation rows across 13 held-out recordings, from four trainees in three experience tiers (junior pre-resident, senior pre-resident, surgical trainee), three glove colors, and 13 surgical training tasks on synthetic-skin pads.

\textbf{What makes the release unique.} Each frame includes the multi-sensor sensor stack (camera intrinsics, MPS hand landmarks with per-hand confidence, timestamps, and eye gaze, with SLAM extrinsics for the recordings where SLAM succeeded). Evaluation labels are initialized from Aria Machine Perception Services (MPS), whose visual-inertial SLAM, multi-camera stereo, and online calibration are unavailable to monocular RGB models, so the reference comes from analytics that the evaluated models cannot access. The labels are sensor-fused, human-validated pseudo-ground truth: every evaluation frame passes primary review, with examiner review and surgeon adjudication of flagged cases. We quantify human correction and inter-annotator agreement (\S\ref{sec:iaa}), characterize label-model mismatch (\S\ref{sec:labelquality}), and validate the labels against an external multi-view triangulated reference (\S\ref{sec:extval}).

\textbf{What the benchmark reveals.} We evaluate eight methods under zero-shot transfer: six MANO-based~\cite{romero2017mano} reconstructors, the two-hand IntagHand, and the keypoint-only MediaPipe Hands (\S\ref{sec:baselines}). Three findings stand out. (i)~ Backbones pretrained using egocentric data are associated with stronger zero-shot transfer, and the ego-pretrained JointTransformer is best at 14.95\,mm PA-MPJPE. (ii)~A substantial surgical domain gap remains, and the widely deployed MediaPipe Hands detects a hand on only 24.1\% of evaluation rows. (iii)~Aligned-pose accuracy is inadequate; most backbones do not predict where the hand is in the camera frame, which AR overlay and robot learning from demonstration require (\S\ref{sec:discussion}).

As a lightweight diagnostic baseline, we developed \textbf{JRN} (Joint-Residual Network), a 58K-parameter two-headed (pose + translation) residual corrector trained on the PR-disjoint training split with the backbone frozen. Its gains persist under a leave-one-wearer-out protocol (10.96-14.85\,mm PA-MPJPE for the two ego-pretrained backbones) and reach the benchmark's empirical label resolution scale of about one centimeter.  Methods are compared on the zero-shot table, whose 15 to 30\,mm spread lies well above that scale. Its translation head supplies the camera-frame root that most backbones lack, and measuring the remaining placement and fingertip error is a central purpose of the benchmark.

\begin{findingbox}{Contributions}
\textbf{Dataset.} EgoSurgHands, a benchmark for real-captured egocentric hand pose estimation in open surgery training on synthetic pads.\par
\textbf{Labels.} Sensor-fused, human-validated pseudo-ground truth initialized from a multi-sensor stack, with a quantified annotation audit, a label-model mismatch diagnostic, and an external multi-view triangulated validation.\par
\textbf{Benchmark.} Eight methods evaluated zero-shot on five metrics with 95\% PR-bootstrap CIs and a leave-one-wearer-out protocol; ego-pretrained models transfer best, with 18 to 40\% lower PA-MPJPE than the non-egocentric reference.\par
\textbf{Diagnostic baseline.} JRN, a 58K-parameter residual corrector that attaches to any frozen backbone and reports pose and placement corrections separately.\par
\textbf{Release.} The dataset, with datasheet and Croissant metadata, and reproducible benchmark code at \url{https://huggingface.co/datasets/surgicalinformatics/EgoSurgHands} and \url{https://github.com/monk1337/egoSurgHands}.
\end{findingbox}

\section{Related Work}
\label{sec:related}

\paragraph{Egocentric hand-pose datasets.} Generalist egocentric corpora cover skilled and everyday activities (EgoExo4D~\cite{egoexo4d}, ARCTIC~\cite{arctic}, AssemblyHands~\cite{assemblyhands}, HOT3D~\cite{hot3d}; sizes in Tab.~\ref{tab:datasets}). Recent large-scale captures include GigaHands~\cite{fu2025gigahands} (51-camera markerless rig), SHOW3D~\cite{rim2026show3d} (in-the-wild hand-object interaction), and EgoDex~\cite{hoque2026egodex} (egocentric manipulation with on-device 3D hand tracking), alongside smaller hand-object (DexYCB~\cite{dexycb}, HO-3D~\cite{ho3d}, HOI4D~\cite{hoi4d}) and multi-view studio corpora (FreiHAND~\cite{freihand}, HanCo~\cite{hanco}, InterHand2.6M~\cite{interhand}). None target surgical training; the closest, HOT3D's office scenario, lacks surgical instruments, glove appearance, and fine-motor grip patterns.

\paragraph{Surgical and OR datasets.} No surgical dataset offers real egocentric articulated 3D hand pose. Skill-assessment sets (JIGSAWS~\cite{jigsaws}, AIxSuture~\cite{aixsuture}) bundle video and rater scores without hand labels, and SutureBot~\cite{suturebot} ships dVRK robot demonstrations; EgoSurgery-Phase~\cite{egosurgeryphase} and EgoSurgery-Tool~\cite{egosurgerytool} use head-mounted Tobii cameras but label only phase, tools, and 2D hand boxes at low fps; the multimodal OR sets 4D-OR~\cite{fourdor} and MM-OR~\cite{mmor} carry no hand labels, and EgoExOR~\cite{egoexor} only wrist and palm tracking (84K frames over two spine procedures); and the egocentric surgical hand-pose benchmark POV-Surgery~\cite{povsurgery} is synthetic (88K frames, three orthopedic tools). The multi-view benchmark of \citet{fischer2026multiview} (68K frames, 3K manually annotated 2D poses with triangulated 3D, replica operating room) is complementary: EgoSurgHands adds the surgeon's own head-worn view, open surgery training tasks, the full multi-sensor sensor stack, and sensor-fused, human-validated labels.

\paragraph{Hand-pose backbones and pseudo-GT precedent.} Our panel (\S\ref{sec:baselines}) spans the dominant MANO-regression paradigms, two-hand reconstruction through IntagHand~\cite{li2022intaghand}, which models both interacting hands jointly because surgical work is predominantly bimanual, and the keypoint-only MediaPipe Hands~\cite{mediapipe}. Further reconstructors such as FrankMocap~\cite{frankmocap} and Hamba~\cite{hamba} can be added through the same 21-joint wrapper interface (Appx.~\ref{appx:backbones}). Labels obtained automatically or from additional sensors, rather than by per-frame manual annotation, are standard practice (motion capture in ARCTIC/HOT3D; multi-view MANO fitting in FreiHAND/HanCo; robot joint states recorded during ALOHA~\cite{alohaopen} teleoperation; ASR- and LLM-derived captions in Quilt-1M~\cite{quilt}); we use Aria MPS (visual-inertial SLAM + stereo + online calibration), unavailable to monocular RGB models at inference. We follow Datasheet~\cite{datasheets} and Croissant~\cite{croissant} documentation standards (eight RAI fields, Appx.~\ref{app:rai}); incidental-face handling follows EgoBlur~\cite{egoblur}.

\section{EgoSurgHands Dataset}
\label{sec:dataset}

\subsection{Recording Setup and Hardware}
\label{sec:setup}

EgoSurgHands was captured using the Meta Project Aria research-edition headset~\cite{aria} (synchronized RGB, two monochrome SLAM cameras, eye tracking, audio, IMU). The forward-facing RGB camera records 1408$\times$1408 video through a fisheye lens (FisheyeRadTanThinPrism model); 2D labels and the PA-P2D metric use a pinhole approximation ($f{=}606.5$, $c{=}704$). Capture rate 30\,fps, with frames released at 5\,Hz (every 6th frame); sensor-profile details and Aria stack diagram in Appx.~Fig.~\ref{fig:aria-device}. Each session is submitted to Meta MPS, which returns SLAM trajectory, 21-joint hand landmarks at 30\,Hz, eye-gaze direction, and a semi-dense scene point cloud. Recordings are on synthetic-skin surgical training pads (no patients, no PHI). Privacy treatment is described in Appx.~\ref{app:rai}.

\begin{figure}[h]
  \centering
  \includegraphics[width=0.9\linewidth]{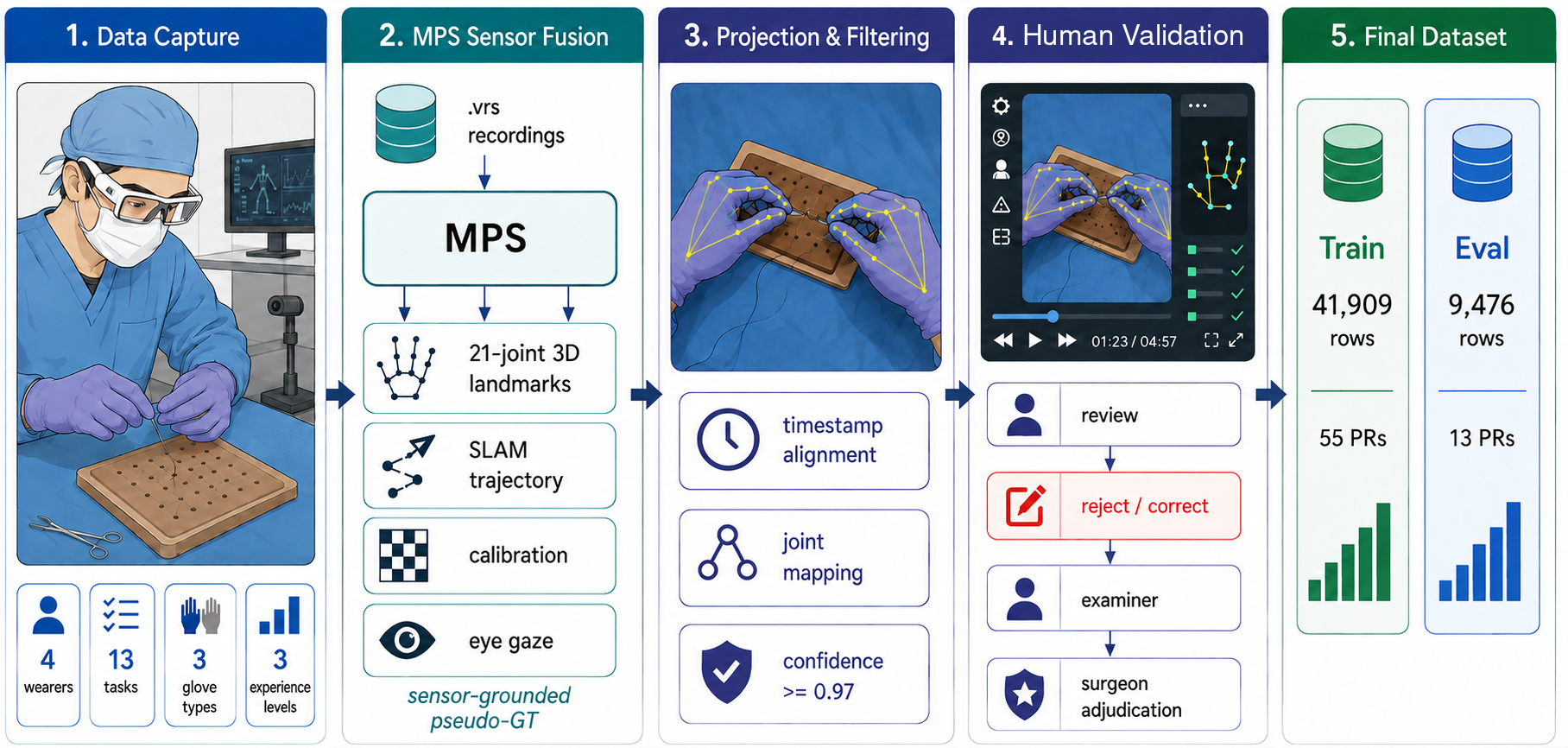}
  \caption{\textbf{EgoSurgHands data pipeline.} \emph{(1) Capture:} four wearers in three experience tiers perform 13 tasks in three glove colors with smart glasses. \emph{(2) MPS:} each \texttt{.vrs} session is processed by Meta Machine Perception Services for SLAM trajectory, 3D hand landmarks, calibration, confidence, and eye gaze; audio is not released. \emph{(3) Pseudo-GT:} landmarks are projected into the RGB frames, time-aligned, confidence-gated at ${\geq}0.97$, and inspected. \emph{(4) Validation:} training rows use gated MPS labels (41{,}909 rows / 55 PRs); evaluation rows are further validated by multi-rater review, examiner review, and surgeon adjudication (9{,}476 rows / 13 held-out PRs).}
  \label{fig:pipeline}
\end{figure}

\subsection{Procedures and Participants}
\label{sec:tasks}

EgoSurgHands spans four wearers (P1-P4) covering three experience tiers (junior pre-resident: P2; senior pre-resident: P3; resident trainee: P1, P4). P2 is left-hand dominant and the others right-hand dominant, providing hand-side diversity uncommon in egocentric hand-pose datasets. The 68 procedure recordings (55 train / 13 val) cover thirteen distinct tasks grouped into eight task families: simple interrupted, figure-of-eight, horizontal mattress, vertical mattress, two-hand tie, running subcuticular closure, needle loading, and cutting. Tasks are performed on synthetic-skin training pads with a standard suturing kit (needle drivers, toothed forceps, Metzenbaum and suture scissors, vicryl/silk/nylon sutures); three glove colors appear (blue, white, brown). All 661 brown-glove evaluation hand instances come from wearer P4, so the P4 fold of the leave-one-wearer-out protocol doubles as a leave-all-brown-glove test, and we treat this stratum as wearer-confounded (\S\ref{sec:per-condition}, \S\ref{sec:lowo}).

\subsection{Ground Truth Pipeline}
\label{sec:pseudo-gt}

Pseudo-GT is derived from Aria Machine Perception Services (MPS), Meta's sensor-fusion pipeline that combines visual-inertial SLAM, stereo, and online calibration to produce 21 3D hand landmarks at 30\,Hz with per-hand tracking confidence. Because MPS uses inertial and stereo cues unavailable to vision-only models, its labels come from a fundamentally different sensing process than any tested baseline. Device-frame landmarks are transformed into the RGB camera frame using the MPS online calibration of the same recording session (\texttt{online\_calibration.jsonl}) and an MPS-to-MANO joint permutation, with per-session temporal re-pairing; we keep a hand only if its MPS tracking confidence is $\geq 0.97$ at both bracketing MPS samples. Projection correctness is verified by manual inspection across all evaluation sessions. We refer to the resulting evaluation labels as \emph{sensor-fused, human-validated pseudo-ground truth}. Because these labels are pseudo-ground truth rather than motion capture, residual benchmark error combines model error, MPS uncertainty, calibration effects, and MPS-to-MANO skeletal-representation differences; we quantify these in \S\ref{sec:iaa} and \S\ref{sec:labelquality} and validate the labels against an external reference in \S\ref{sec:extval}.

\subsection{Inter-Annotator-Agreement Validation}
\label{sec:iaa}

A custom multi-rater annotation tool (see Appx.~\ref{appx:annotator}) overlays the MPS-projected 21-joint skeleton on the RGB crop and supports keep / reject / drag-correct actions, so reviewers remove or fix frames with projection drift, partial occlusion, motion blur, or hand-side errors. Expert annotators perform primary review, an examiner pass cross-reviews flagged cases, a surgeon adjudicates residual disagreements as final, and a three-rater audit subset measures agreement (below; full procedure and statistics in Appx.~\ref{app:iaa}). Frames passing both the $\geq 0.97$ MPS gate and the human-review gate form the human-validated evaluation split (9{,}476 hand instances, 13 PRs).

\paragraph{Correction, rejection, and agreement.} Across 9{,}476 review decisions, 43.1\% of frames were accepted as-is and 51.9\% after a drag correction (median 8.2\,mm; mean 9.8\,mm over the 4{,}914 corrections; 5.3\,mm over all reviewed hand instances); only 5.1\% were rejected. An offset is recorded only after an annotator flags an overlay and is stored as a global image-plane translation, so the 51.9\% reflects how often annotators made a sub-centimeter adjustment rather than how often a label failed, and these values quantify the magnitude of human intervention, not per-joint 3D error. On a 500-frame stratified audit subset in which three annotators independently made keep/reject decisions, pooled pairwise agreement is 72.9\%. Surgeon adjudication resolved every escalated frame and changed the label on 55.5\% of them, as expected for the hardest cases, which are the ones selected for escalation.

\subsection{Splits and Held-Out Settings}
\label{sec:splits}

EgoSurgHands includes two PR-disjoint splits: a 41{,}909-row training partition (55 PRs) and a 9{,}476-row human-validated evaluation partition (13 PRs). Both splits share the same schema (RGB image, per-frame intrinsics and extrinsics, MPS landmarks, hand bbox, hand-side, and per-frame metadata: recording day, wearer, experience tier, handedness, glove color, task). The 13 evaluation PRs span all four wearers, all three experience tiers, all three glove colors, and a balanced hand-side distribution. Any adapter (e.g., JRN, \S\ref{sec:protocols}) is trained on the training partition and applied to the evaluation partition; no evaluation frame is used for training.

\subsection{Statistics and Comparison}
\label{sec:stats}

The evaluation partition is jointly stratified by wearer (P1: 4{,}694, P2: 3{,}198, P3: 923, P4: 661 hand instances), experience tier, glove color (blue 4{,}694, white 4{,}121, brown 661), and task family, with approximately even hand sides; per-PR row counts range from 75 to 2{,}162 (full breakdown in Appx.~\ref{app:stats}).

Table~\ref{tab:datasets} positions EgoSurgHands among twelve widely used 3D hand-pose corpora across dimensions relevant to surgical applications: capture device, hardware time-synchronization, frame count, surgical-task count, and hand-annotation depth. EgoSurgHands is the only entry that combines surgical training scope with Aria capture, hardware-synchronized sensor fusion, and 21-joint MPS pseudo-GT. It is the only egocentric corpus focused on surgical training tasks at a sensor-anchored quality bar and, among the surgical entries, the only real egocentric capture with human-validated 3D hand-pose labels.

\begin{table*}[t]
  \centering
  \scriptsize
  \setlength{\tabcolsep}{3pt}
  \caption{\textbf{Where EgoSurgHands sits among ego and exo hand-pose datasets.} \textbf{Surg.}: real-captured surgical activity; \textbf{HW sync}: hardware time-synchronized cameras; \textbf{Frames}: as reported by each paper (multi-view sets may count every camera image); \textbf{Artic.\ 3D}: articulated 3D hand joints; n/r = not reported; n/a = single camera. EgoSurgHands is the only entry that is egocentric, real-captured surgical, and annotated with articulated 3D hands, and it adds hardware-synchronized Aria sensor fusion with MPS pseudo-GT.}
  \label{tab:datasets}
  \renewcommand{\arraystretch}{0.9}
  \resizebox{\textwidth}{!}{%
  \begin{tabular}{llcclcrrlc}
    \toprule
    Dataset & Domain & Surg. & Ego & Device & HW sync & Frames & \#Tasks & Hand annotations & Artic.\ 3D \\
    \midrule
    EgoExo4D & Skilled activity & \xmark & \cmark & Aria+exo & \xmark & 1{,}286 h & 43 & 21 kp & \cmark \\
    ARCTIC & Object manip. & \xmark & \cmark & Helmet+exo & \cmark & 2.1M & 11 & full MANO & \cmark \\
    AssemblyHands & Toy assembly & \xmark & \cmark & Headset+exo & n/r & 490K & 1 & 21 kp & \cmark \\
    HOT3D & Object int. & \xmark & \cmark & Aria+Quest3 & \cmark & 3.7M & 4 & UmeTrack + MANO & \cmark \\
    FreiHAND & Generic & \xmark & \xmark & Multi-view rig & \cmark & 130K & 1 & full MANO & \cmark \\
    POV-Surgery & Surgical (synth.) & \xmark & \cmark & Sim cam & n/a & 88K & 3 & full MANO & \cmark \\
    AIxSuture & Suturing & \cmark & \xmark & GoPro Hero 5 & n/a & $\sim$2.8M & 1 & none & \xmark \\
    EgoExOR & OR multi-modal & \cmark & \cmark & Aria+Kinect & \xmark & 84K & 2 & wrist/palm tracking & \xmark \\
    GigaHands & Bimanual activities & \xmark & \xmark & 51-camera rig & \xmark & 183M & 191 & auto 3D & \cmark \\
    SHOW3D & Hand-object (wild) & \xmark & \cmark & Quest3+back rig & \cmark & 4.3M & n/r & 3D kp + MANO & \cmark \\
    EgoDex & Tabletop manip. & \xmark & \cmark & Vision Pro & n/r & 90M & 194 & device tracking & \cmark \\
    Fischer et al. & Surgical (replica OR) & \cmark & \xmark & Multi-view rig & \cmark & 68K & 2 & 3K 2D + tri.\ 3D & \cmark \\
    \midrule
    \rowcolor{green!10}
    \textbf{EgoSurgHands (ours)} & \textbf{Surgical training} & \cmark & \cmark & \textbf{Aria} & \cmark & \textbf{42K + 9K} & \textbf{13} & \textbf{21 jt (MPS)} & \cmark \\
    \bottomrule
  \end{tabular}}
\end{table*}

Per-task hand-mesh fingerprints (Appx.~Fig.~\ref{fig:fingerprints}) show that EgoSurgHands captures task-discriminative motion patterns at the hand-mesh level.

\section{Benchmark Protocols}
\label{sec:protocols}

\subsection{Baseline Models}
\label{sec:baselines}

We benchmark eight hand-pose methods spanning three modeling families and ship one diagnostic adaptation baseline, the Joint-Residual Network (JRN). The six MANO-based backbones are HaMeR~\cite{hamer} (our non-egocentric reference), WiLoR~\cite{wilor}, WildHands~\cite{wildhands} (the closest publicly available ego-trained model), JointTransformer~\cite{jt} (a second ego-pretrained backbone with a different architecture, trained on ARCTIC), MeshGraphormer~\cite{meshgraphormer}, and SimpleHand~\cite{simplehand} (a minimum-capacity baseline). IntagHand~\cite{li2022intaghand} reconstructs two interacting hands jointly, addressing the bimanual character of surgical scenes, and MediaPipe Hands~\cite{mediapipe} is keypoint-only; it detects a hand on 24.1\% of evaluation rows, so its errors are computed on detected rows only. All checkpoints are the original authors' public releases, and no Aria pad data appears in any backbone's pretraining mixture; architectures and pretraining corpora are detailed in Appx.~\ref{appx:backbones}. All methods are evaluated zero-shot on the human-validated EgoSurgHands evaluation partition (9{,}476 hand instances); all eight are additionally evaluated with the JRN head attached (trained on the dedicated training partition of 41{,}909 hand instances).

\paragraph{JRN: a diagnostic adaptation baseline.} JRN is a 58K-parameter two-headed MLP attached on top of any frozen backbone and shipped with the benchmark as a diagnostic baseline. It is trained on the dedicated training partition with MPS supervision and evaluated against MPS-derived labels. It is applied to the held-out evaluation partition and retrained per fold in the leave-one-wearer-out protocol (\S\ref{sec:protocol}).

\paragraph{Formal definition and training.} Let $f$ be the frozen backbone and $x = f(I) \in \mathbb{R}^{21\times 3}$ its prediction with wrist $x_0$. JRN computes $\hat{x} = x + \delta_p + \delta_t \cdot \mathbf{1}_{21}$, where the pose head produces $\delta_p \in \mathbb{R}^{21\times 3}$ from $x - x_0$ (32{,}831 params) and the translation head produces $\delta_t \in \mathbb{R}^3$ from $[\,x - x_0,\, x_0\,]$ (25{,}475 params); full module code, per-layer breakdown, and the diagram are in Appx.~\ref{appx:jrn} to \ref{appx:jrn-diagram}. JRN is trained on the 41{,}909 training hand instances with MPS supervision: pose target $Y_p = (x_{\mathrm{GT}} - x_{\mathrm{GT},0}) - (x - x_0)$, translation target $Y_t = x_{\mathrm{GT},0} - x_0$, loss $\mathcal{L} = \mathrm{MSE}(\delta_p, Y_p) + \mathrm{MSE}(\delta_t, Y_t)$, AdamW~\cite{loshchilov2019adamw} $\mathrm{lr}{=}10^{-3}$, $\mathrm{wd}{=}10^{-4}$, batch $512$, 60 epochs, seed $42$ (Appx.~\ref{appx:training}). Both heads are zero-initialized, so JRN matches the ZS baseline at $t{=}0$; after training, JRN improves all five metrics on all eight methods (Tab.~\ref{tab:headline}). The translation head is invariant under Procrustes alignment, so it leaves PA-MPJPE / MPJPE-RR / PA-P2D bit-identical; the pose head leaves Root err approximately unchanged because $Y_p[0] \equiv 0$ by construction. MPJPE-abs is the one metric that depends on both heads (Appx.~\ref{appx:orthogonal}). The pose head measures how much root-relative configuration can be corrected; the translation head supplies the camera-frame placement the backbones do not predict. We report these effects separately rather than combining them into a single claim about accuracy.

\begin{figure}[t]
\centering
\resizebox{\linewidth}{!}{%
\begin{tikzpicture}[
    >=Latex,
    every node/.style={font=\footnotesize},
    box/.style={draw, thick, rounded corners=3pt, align=center, inner sep=3pt, minimum height=0.95cm},
    img/.style={box, fill=yellow!12, minimum width=1.7cm},
    bb/.style={box, fill=gray!18, minimum width=2.5cm},
    pred/.style={box, fill=white, very thick, minimum width=2.0cm},
    encP/.style={box, fill=blue!8, draw=blue!55, minimum width=2.45cm},
    encT/.style={box, fill=orange!8, draw=orange!55, minimum width=2.45cm},
    mlpP/.style={box, fill=blue!12, draw=blue!60, minimum width=2.1cm},
    mlpT/.style={box, fill=orange!12, draw=orange!60, minimum width=2.1cm},
    headP/.style={box, fill=blue!18, draw=blue!70, minimum width=1.7cm},
    headT/.style={box, fill=orange!18, draw=orange!70, minimum width=1.7cm},
    deltaP/.style={font=\small\bfseries, color=blue!70!black},
    deltaT/.style={font=\small\bfseries, color=orange!75!black},
    sumc/.style={circle, draw, very thick, minimum size=0.8cm, inner sep=0pt, fill=green!22, font=\Large\bfseries},
    outbox/.style={box, fill=green!18, very thick, minimum width=2.3cm, font=\footnotesize\bfseries},
    arr/.style={->, very thick},
    arrP/.style={->, very thick, blue!65},
    arrT/.style={->, very thick, orange!75},
    edgelbl/.style={font=\scriptsize, inner sep=1.5pt, fill=white, rounded corners=1pt}
]
\node[img]  (img)  at (0, 0)    {Hand crop\\ \scriptsize RGB image $I$};
\node[bb]   (bb)   at (2.55, 0) {\textbf{Frozen backbone $f$}\\ \scriptsize any of the 8 methods};
\node[pred] (pred) at (5.25, 0) {\textbf{pred}\\ $x \in \mathbb{R}^{21\times 3}$};
\draw[arr] (img) -- (bb);
\draw[arr] (bb) -- (pred);
\node[encP]  (rrel)  at (7.75, 0.8)  {\textbf{root-rel.}\\ $x{-}x_0 \in \mathbb{R}^{63}$};
\node[mlpP]  (poseT) at (10.1, 0.8)  {\textbf{Pose trunk}\\ \scriptsize $63{\to}128{\to}128$};
\node[headP] (poseH) at (12.2, 0.8)  {\textbf{Pose head}\\ \scriptsize $128{\to}63$};
\node[deltaP] (dP)   at (13.75, 0.8) {$\delta_{p}$};
\node[encT]  (rmix)  at (7.75, -0.8) {\textbf{root-rel.\,$\oplus$\,root}\\ $[x{-}x_0, x_0] \in \mathbb{R}^{66}$};
\node[mlpT]  (transT) at (10.1, -0.8) {\textbf{Trans trunk}\\ \scriptsize $66{\to}128{\to}128$};
\node[headT] (transH) at (12.2, -0.8) {\textbf{Trans head}\\ \scriptsize $128{\to}3$};
\node[deltaT] (dT)    at (13.75, -0.8) {$\delta_{t}$};
\draw[arrP] (pred.north east) to[out=0, in=180] (rrel.west);
\draw[arrT] (pred.south east) to[out=0, in=180] (rmix.west);
\draw[arrP] (rrel) -- (poseT);  \draw[arrP] (poseT) -- (poseH);  \draw[arrP] (poseH) -- (dP);
\draw[arrT] (rmix) -- (transT); \draw[arrT] (transT) -- (transH); \draw[arrT] (transH) -- (dT);
\begin{scope}[on background layer]
\node[fit=(rrel)(poseT)(poseH)(dP), draw=blue!50, dashed, very thick, rounded corners=6pt, inner sep=4pt] (poseGroup) {};
\node[fit=(rmix)(transT)(transH)(dT), draw=orange!60, dashed, very thick, rounded corners=6pt, inner sep=4pt] (transGroup) {};
\end{scope}
\node[anchor=south, font=\scriptsize\bfseries\color{blue!70!black}, inner sep=1.5pt] at (poseGroup.north) {Pose head (32{,}831 params, zero-init)};
\node[anchor=north, font=\scriptsize\bfseries\color{orange!75!black}, inner sep=1.5pt] at (transGroup.south) {Translation head (25{,}475 params, zero-init)};
\node[sumc] (sum) at (15.25, 0) {$+$};
\draw[arrP] (dP.east) to[out=0, in=110] (sum.north);
\draw[arrT] (dT.east) to[out=0, in=-110] (sum.south);
\draw[arr, dashed] (pred.north) to[out=90, in=180] ++(1.1,1.5) -- node[edgelbl, pos=0.5, above] {skip: $x$} ++(6.4,0) to[out=0, in=135] (sum.north west);
\node[outbox] (out) at (17.35, 0) {corrected\\ $\hat{x} = x + \delta_{p} + \delta_{t}\mathbf{1}_{21}$};
\draw[arr] (sum) -- (out);
\end{tikzpicture}%
}
\caption{\textbf{JRN architecture.} A frozen backbone predicts 21 camera-frame joints $x$; the pose head (blue) corrects the root-relative pose and the translation head (orange) the wrist position. Both are zero-initialized (identity at start); details in Appx.~\ref{appx:jrn-diagram}.}
\label{fig:jrn-compact}
\end{figure}
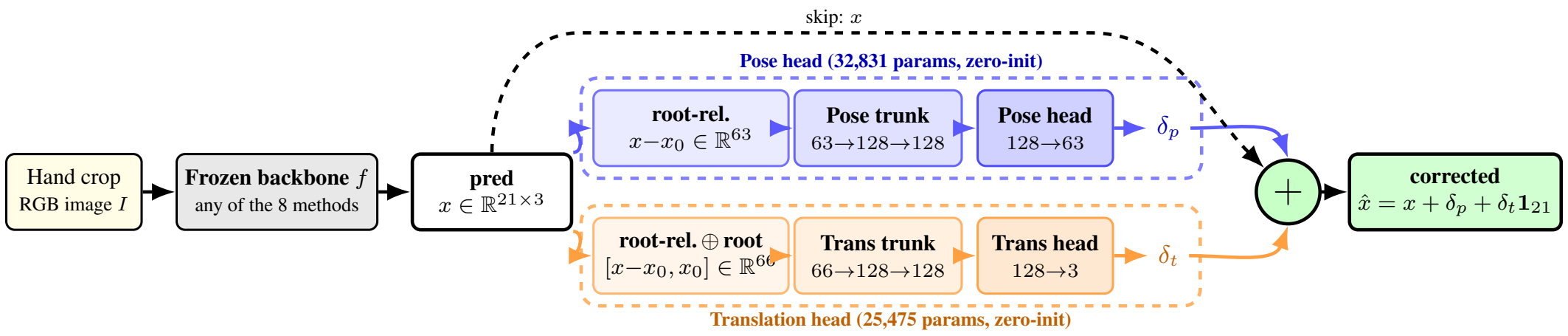

\subsection{Evaluation Metrics}
\label{sec:metrics}

We report five error metrics against the sensor-fused, human-validated Aria MPS labels, all on the OP\_VALID 20-joint subset (dropping the OpenPose thumb-CMC joint, which has no Aria MPS analog, \S\ref{sec:pseudo-gt}). \textbf{PA-MPJPE-MPS (mm)}, the metric and standard in 3D and egocentric hand pose~\cite{hamer,egoexo4d}: per-joint L2 to MPS after per-frame Umeyama Procrustes~\cite{umeyama1991} (rotation + uniform scale + translation). \textbf{MPJPE-RR (mm)}: root-relative MPJPE (both wrists at the origin; retains scale and rotation). \textbf{Root err (mm)}: wrist-translation error. \textbf{MPJPE-abs (mm)}: unaligned MPJPE, which depends on both pose and global placement and matters for AR overlay. \textbf{PA-P2D (px)}: Procrustes-aligned 3D projected through the Aria pinhole approximation ($f{=}606.5$, $c{=}704$), a per-joint image-space check of PA-MPJPE. Every number carries a 95\% bootstrap CI from 1{,}000 resamples of procedure recordings rather than frames, a conservative choice because frames within a PR are temporally correlated; the intervals describe uncertainty over the 13 evaluation PRs.

\subsection{Experimental Protocol}
\label{sec:protocol}

Train and evaluation partitions are PR-disjoint by construction (55 train / 13 evaluation PRs). They share wearers and protocols, so we call this the within-corpus split and add a leave-one-wearer-out protocol that holds out each wearer in turn. The benchmark trains JRN once on all 41{,}909 training hand instances and evaluates on the 9{,}476 evaluation hand instances, with metrics weighted by per-PR instance count.

\paragraph{Leave-one-wearer-out.} For each wearer, JRN is retrained on training-partition rows from the other three wearers and evaluated on the held-out wearer (Tab.~\ref{tab:lowo}). The pretrained backbones remain zero-shot throughout and never train on EgoSurgHands, so the zero-shot comparison is protected from train-test-overlap explanations by construction.

\section{Experiments and Results}
\label{sec:experiments}

\subsection{Zero-Shot Transfer and JRN Correction}
\label{sec:headline}

Table~\ref{tab:headline} reports zero-shot performance and JRN-corrected performance under the standard ML protocol: \textbf{train the JRN head on the dedicated training partition (41{,}909 hand instances, 55 PRs) and evaluate on the held-out human-validated evaluation partition (9{,}476 hand instances, 13 PRs).}

\begin{figure}[t]
  \centering
  \includegraphics[width=0.92\linewidth]{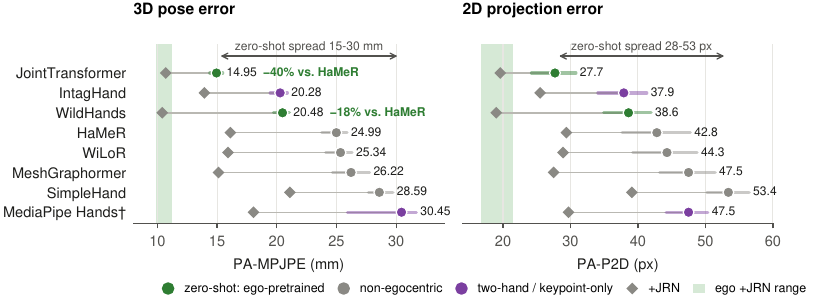}
  \caption{\textbf{Benchmark: zero-shot error with 95\% PR-bootstrap CIs, and the effect of JRN.} Ego-pretrained backbones (green) have the lowest zero-shot error among single-hand methods, 40\% (JointTransformer) and 18\% (WildHands) below the non-egocentric reference HaMeR. Left: PA-MPJPE (mm, vs Aria MPS). Right: PA-P2D (px); protocol as in Tab.~\ref{tab:headline}. Circles: zero-shot, colored by model family, with CI bars; diamonds: +JRN. The green band spans the two ego-pretrained +JRN results, which reach the label resolution scale (\S\ref{sec:labelquality}) and are not ranked. $^\dagger$MediaPipe on the 24.1\% of rows where it detects a hand.}
  \label{fig:forest_combined}
\end{figure}

\begin{findingbox}{Ego-pretrained models transfer best zero-shot to egocentric surgical training video}
Evaluated zero-shot on egocentric surgical training video, the two backbones pretrained with egocentric data have the lowest errors among the single-hand models in our panel: JointTransformer reaches 14.95\,mm [14.39, 15.46] and WildHands-ZS 20.48\,mm [19.72, 21.06], 40\% and 18\% below the non-egocentric reference HaMeR-ZS at 24.99\,mm [23.77, 25.89]. Their 95\% CIs lie below those of all four non-egocentric MANO backbones (24.99 to 28.59\,mm), and both are also lower on the 2D metric (27.7 and 38.6 vs.\ 42.8 to 53.4\,px PA-P2D) and on every held-out wearer (\S\ref{sec:lowo}). Because architecture, capacity, and pretraining corpus also differ among these backbones, we report this as an association rather than an isolated causal effect of egocentric pretraining.
\end{findingbox}
 The two-hand IntagHand reaches 20.28\,mm, and the keypoint-only MediaPipe Hands reaches 30.45\,mm on the 24.1\% of rows where it detects a hand. Figure~\ref{fig:forest_combined} visualizes the distribution.

After JRN correction trained on the training partition, WildHands+JRN reaches 10.40\,mm [9.91, 10.94] (a 49.2\% reduction from WildHands-ZS) and JointTransformer+JRN reaches 10.70\,mm [10.21, 11.19] (a 28.4\% reduction) on the within-corpus split. These corrected errors reach the benchmark's empirical label resolution scale (\S\ref{sec:labelquality}), so the 0.30\,mm gap between the two chains, like other differences among corrected backbones, lies below what the current labels resolve. Methods are therefore compared on the zero-shot table, whose 15 to 30\,mm spread lies well above that scale, with JRN as a diagnostic baseline.

JRN improves every one of the 13 evaluation PRs for all six backbones, with no regressions (Fig.~\ref{fig:scatter}).

\begin{table*}[t]
  \centering
  \footnotesize
  \setlength{\tabcolsep}{4pt}
  \caption{\textbf{Benchmark on the evaluation partition (9{,}476 hand instances, 13 PRs), all five metrics.} JRN is the 58K-parameter pose + translation corrector trained on the PR-disjoint training partition (41{,}909 hand instances, 55 PRs) with the same script and configuration for every method. Errors in mm vs Aria MPS landmarks; PA-P2D in pixels through the Aria pinhole approximation ($f{=}606.5$, $c{=}704$). All metrics on the OP\_VALID 20-joint subset; PA metrics use Umeyama Procrustes. Best per column in bold. +JRN values reach the label resolution scale (\S\ref{sec:labelquality}) and are not ranked. $^\dagger$MediaPipe errors are on the 24.1\% of evaluation rows where it detects a hand (2{,}283 of 9{,}476); its JRN is trained on the 11{,}349 training rows where it detects one.}
  \label{tab:headline}
  \resizebox{\linewidth}{!}{%
  \begin{tabular}{lrrrrrrrrrr}
    \toprule
     & \multicolumn{2}{c}{PA-MPJPE (mm)} & \multicolumn{2}{c}{MPJPE-RR (mm)} & \multicolumn{2}{c}{Root err (mm)} & \multicolumn{2}{c}{MPJPE-abs (mm)} & \multicolumn{2}{c}{PA-P2D (px)} \\
    \cmidrule(lr){2-3}\cmidrule(lr){4-5}\cmidrule(lr){6-7}\cmidrule(lr){8-9}\cmidrule(lr){10-11}
    Backbone & ZS & +JRN & ZS & +JRN & ZS & +JRN & ZS & +JRN & ZS & +JRN \\
    \midrule
    \rowcolor{green!10}
    \multicolumn{11}{l}{\textit{\textbf{Ego-pretrained backbones}}} \\
    \rowcolor{green!10}
    JointTransformer~\cite{jt} & \textbf{14.95} & 10.70 & 133.7 & 40.8 & 355.5 & 75.4 & 352.7 & 79.0 & 27.7 & 19.6 \\
    \rowcolor{green!10}
    WildHands~\cite{wildhands} & 20.48 & \textbf{10.40} & 133.3 & \textbf{40.1} & 550.4 & \textbf{73.8} & 533.3 & \textbf{75.8} & 38.6 & \textbf{19.0} \\
    \midrule
    \multicolumn{11}{l}{\textit{\textbf{Non-egocentric backbones}}} \\
    HaMeR~\cite{hamer} & 24.99 & 16.11 & 157.7 & 50.7 & 313.8 & 105.6 & 365.0 & 91.7 & 42.8 & 29.4 \\
    MeshGraphormer~\cite{meshgraphormer} & 26.22 & 15.12 & 151.4 & 51.9 & 344.9 & 106.1 & 400.5 & 96.1 & 47.5 & 27.5 \\
    WiLoR~\cite{wilor} & 25.34 & 15.91 & 151.4 & 51.1 & 314.0 & 102.8 & 369.9 & 88.0 & 44.3 & 28.9 \\
    SimpleHand~\cite{simplehand} & 28.59 & 21.10 & 137.8 & 69.5 & 324.0 & 141.4 & 356.1 & 109.8 & 53.4 & 39.1 \\
    \midrule
    \multicolumn{11}{l}{\textit{\textbf{Two-hand and keypoint-only methods}}} \\
    IntagHand~\cite{li2022intaghand} & 20.28 & 13.94 & 157.9 & 50.8 & 322.5 & 92.5 & 363.4 & 90.5 & 37.9 & 25.5 \\
    MediaPipe Hands$^\dagger$~\cite{mediapipe} & 30.45 & 18.05 & 122.9 & 63.3 & 328.6 & 121.4 & 386.1 & 94.9 & 47.5 & 29.7 \\
    \bottomrule
  \end{tabular}%
  }
\end{table*}

\subsection{Per-Condition Analysis}
\label{sec:per-condition}

\begin{figure}[t]
  \centering
  \includegraphics[width=0.92\linewidth]{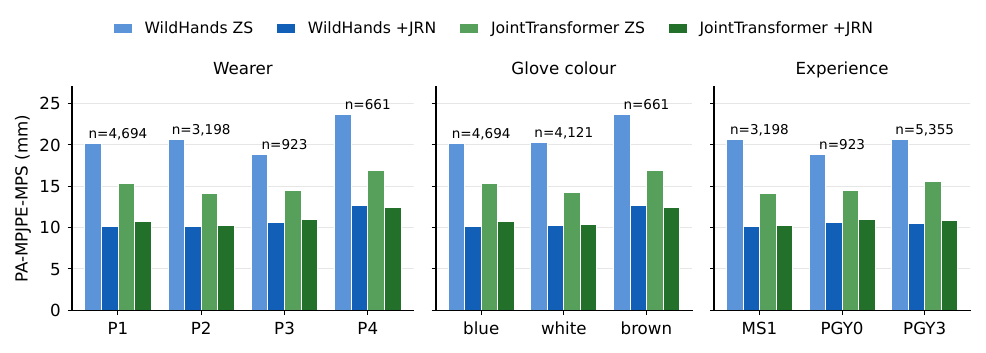}
  \caption{\textbf{Per-stratum PA-MPJPE-MPS for WildHands and JointTransformer (ZS vs +JRN).} JRN lowers error in every stratum; P4 (all brown-glove hand instances, $n{=}661$) is the hardest for both backbones. Glove color, experience tier, and wearer are confounded in the evaluation split, so stratum differences are not attributed to any single factor.}
  \label{fig:strata}
\end{figure}

Per-stratum errors (full table in Appx.~\ref{app:figs}, Tab.~\ref{tab:strata}; visualized in Fig.~\ref{fig:strata}) for WH+JRN/JT+JRN are 10.19-10.97\,mm for P1 ($n{=}4{,}694$), P2 ($n{=}3{,}198$), and P3 ($n{=}923$), and 12.39-12.64\,mm for P4 ($n{=}661$). By glove color, blue ($n{=}4{,}694$, P1 only) gives 10.19/10.71\,mm, white ($n{=}4{,}121$, P2 and P3) 10.28/10.42\,mm, and brown ($n{=}661$, P4 only) 12.64/12.39\,mm. Glove color and wearer are confounded in the evaluation split, so we report the P4/brown stratum as a wearer-confounded observation and do not attribute it to any single factor. PR-level WH+JRN ranges 9.33\,mm (PR115) to 12.64\,mm (PR184, the P4/brown session, $n{=}661$). Per-joint (Appx.~\ref{app:figs}, Fig.~\ref{fig:per_joint_heatmap}): JRN reduces wrist (HaMeR 41.2$\to$20.4\,mm) and MCP joints (index\_MCP 31.4$\to$13.3\,mm) most; MCPs become near-anatomically constrained (6-8\,mm for ego). Fingertips remain hardest across all configurations (16-35\,mm post-JRN). Per-task (Fig.~\ref{fig:per_task}, with per-task $n$ above each bar): WH+JRN 9.33-12.64\,mm; JT+JRN 9.44-13.17\,mm.

\subsection{Ablation Study}
\label{sec:ablation}

Tab.~\ref{tab:ablation} (Appx.~\ref{app:figs}) lists zero-shot and JRN-corrected PA-MPJPE side by side for representative backbones, and Fig.~\ref{fig:delta} (Appx.) visualizes the per-backbone JRN improvement. Post-JRN values sit at or near the label resolution scale (\S\ref{sec:labelquality}), so post-JRN gaps between backbones are not read as evidence about how egocentric pretraining and residual correction combine.

\section{Discussion}
\label{sec:discussion}

The hardest condition is wearer P4's brown-glove sessions ($n{=}661$; 12-13\,mm for ego backbones, higher for non-ego), a wearer-confounded observation. Fig.~\ref{fig:qual-zs-jrn-gt} (Appx.~\ref{app:figs}) shows representative frames across all three glove conditions.

\paragraph{Aligned pose versus global placement.} Zero-shot PA-MPJPE of 15 to 30\,mm supports comparison for gesture, technique, and skill analysis. The benchmark also exposes large Root err: most backbones are root-relative and do not predict the camera-frame hand position needed for precise AR overlay or robot learning from demonstration, and even with JRN's camera-frame root, MPJPE-abs remains 75 to 79\,mm for the ego-pretrained backbones (88 to 110\,mm for the others), too large for these applications. The reduction in WildHands MPJPE-abs from 533.3 to 75.8\,mm is almost entirely explained by the drop in Root err from 550.4 to 73.8\,mm, a systematic offset that is predominantly in-plane rather than in depth; because PA-MPJPE removes translation by Procrustes alignment, this reduction cannot inflate the aligned metrics.

\subsection{Leave-One-Wearer-Out Evaluation}
\label{sec:lowo}

\begin{table}[t]
  \centering
  \footnotesize
  \setlength{\tabcolsep}{4pt}
  \caption{\textbf{Leave-one-wearer-out PA-MPJPE (mm), zero-shot / +JRN.} For each held-out wearer, JRN is retrained using only training-partition rows from the other three wearers; backbones remain zero-shot. Held-out sizes: P1 4{,}694, P2 3{,}198, P3 923, P4 661. The P4 fold is also the leave-all-brown-glove fold.}
  \label{tab:lowo}
  \begin{tabular}{lcccc}
    \toprule
    Backbone & P1 & P2 & P3 & P4/brown \\
    \midrule
    HaMeR & 25.64 / 17.59 & 24.26 / 18.37 & 23.06 / 14.92 & 26.58 / 27.71 \\
    WildHands & 20.21 / 12.00 & 20.68 / 14.25 & 18.89 / 10.96 & 23.64 / 14.68 \\
    JointTransformer & 15.32 / 12.65 & 14.13 / 11.30 & 14.48 / 11.45 & 16.94 / 14.85 \\
    MeshGraphormer & 24.25 / 17.70 & 27.59 / 17.31 & 27.00 / 14.30 & 32.56 / 15.84 \\
    SimpleHand & 27.54 / 21.47 & 29.67 / 22.57 & 27.00 / 21.36 & 33.06 / 21.29 \\
    \bottomrule
  \end{tabular}
\end{table}

Table~\ref{tab:lowo} reports the wearer-disjoint protocol (\S\ref{sec:protocol}). Pooling the four folds reproduces the published zero-shot values exactly, and the zero-shot association with egocentric pretraining holds in every fold. JRN improves the held-out wearer in 19 of 20 backbone-fold cells, and the two egocentric backbones reach 10.96-14.85\,mm without having seen the evaluation wearer, so the gains are not an in-distribution artifact of shared wearers; as dataset-specific MPS-label adaptation, they are measured against MPS labels rather than an independent 3D reference. The exception, HaMeR on P4, is a backbone-specific effect rather than a property of the stratum, and P4 is the hardest zero-shot fold for all five backbones. These results cover five backbones and four wearers; cross-site and cross-protocol folds would better validate these results.

\subsection{Label Quality and Resolution Scale}
\label{sec:labelquality}

A model-consensus diagnostic across independently trained backbone panels puts the consensus-to-MPS displacement at 11.7 to 13.2\,mm, which is stable across panel variants and under leave-one-backbone-out analysis. This may reflect the mismatch between MANO-style and MPS hand representations rather than pure MPS error. The external triangulated validation (\S\ref{sec:extval}) provides a meaningful range for their combined effect. This diagnostic, the human correction signal (\S\ref{sec:iaa}), and the best corrected PA-MPJPE result all fall in the 5 to 13\,mm range, giving the benchmark an empirical resolution scale of about one centimeter. The 15 to 30\,mm zero-shot spread lies well above this scale, so the zero-shot table supports comparing methods; fine-grained separation among corrected models, below about one centimeter, will require labels more precise than the current pseudo-ground truth.

\subsection{External Triangulated Validation}
\label{sec:extval}

To test the MPS labels against an independent reference, four wearers repeated the tasks in all three glove colors while two GoPros and an overhead iPhone, calibrated as one metric rig with a ChArUco board, recorded alongside the Aria glasses (Appx.~\ref{app:extval}). Over 500 frames from 17 recordings, MPS agrees with this reference to a median 9.8\,mm per joint after per-hand alignment and to 15.9\,mm (95\% CI 14.7 to 17.5) after one similarity transform per recording (16.7\,mm with no scale fitted to MPS). Because the reference carries its own keypoint and triangulation error, these values are the upper bound of the MPS error. They support the zero-shot comparisons but not the ranking of corrected models (\S\ref{sec:labelquality}).

\section{Limitations and Conclusion}
\label{sec:conclusion}

\paragraph{Limitations.} The data cover four trainees on synthetic pads, a controlled setting that allows consented, PHI-free capture; live operating rooms and further sites are the natural next step. Labels are sensor-fused, human-validated pseudo-ground truth rather than motion capture and resolve differences of about one centimeter: enough for the 15 to 30\,mm zero-shot comparisons, but not for sub-centimeter differences such as those among JRN-corrected models. The procedure-disjoint split shares wearers (hence the leave-one-wearer-out protocol), JRN is a dataset-specific diagnostic baseline, P4/brown confounds wearer and glove, and intervals cover 13 evaluation PRs.

\paragraph{Conclusion.} EgoSurgHands is, to our knowledge, the first benchmark for real-captured egocentric hand pose in open surgery training on synthetic pads. Within the evaluated panel, egocentric pretraining is associated with stronger zero-shot transfer, a substantial surgical domain gap remains in aligned pose, fingertips, and global placement, and JRN's gains persist on unseen wearers, reaching the roughly 1\,cm label resolution scale. We release the dataset and benchmark code so that EgoSurgHands can serve as a shared testbed for surgical hand understanding, from skill assessment and AR guidance to robot learning from surgeons' demonstrations.

\newpage
\begin{ack}
We thank the lab members who volunteered to wear multi-sensor glasses, and the annotators and surgeons who reviewed, corrected, and adjudicated the evaluation frames. This benchmark is possible because of their time and expertise. We thank the Department of Surgery at Beth Israel Deaconess Medical Center for hosting  recordings and providing the training materials, and the Project Aria team at Meta Reality Labs Research for the Aria Research Kit.

We thank Nebius for GPU compute that powered elements of the zero-shot evaluation of all eight methods, the training of the JRN baselines across every split and wearer fold, and the leave-one-wearer-out study.

We also thank the authors of HaMeR, WiLoR, WildHands, JointTransformer, MeshGraphormer, SimpleHand, IntagHand, and MediaPipe Hands for releasing their models, and the anonymous reviewers and the area chair, whose questions led to the leave-one-wearer-out study and the triangulated validation.

\paragraph{Funding and competing interests.} GPU compute was provided by Nebius and through funding from the Josiah Macy Jr.\ Foundation Artificial Intelligence in Medical Education Grant. The Aria headsets were provided as an in-kind donation from Meta Platforms, Inc. Beyond the support listed above, the authors declare no competing interests.
\end{ack}

\bibliographystyle{plainnat}
\bibliography{references}

\appendix

\section{JRN architecture details}
\label{appx:jrn}

This appendix gives the full PyTorch definition of the Joint-Residual Network (JRN), a per-layer parameter breakdown, the zero-initialization guarantee, and a gradient-flow argument explaining why the pose head and translation head learn orthogonal corrections. A diagram of the same forward pass is given in Appendix~\ref{appx:jrn-diagram}. JRN is dataset-specific MPS-label adaptation.

\subsection{Module definition}
\label{appx:jrn-code}

\begin{small}
\begin{verbatim}
class JRN(nn.Module):
    """Joint-Residual Network. 2-headed MLP residual corrector for monocular
    hand-pose backbones. Pose head learns root-relative joint corrections;
    translation head learns the wrist-position correction. 58,306 params."""
    def __init__(self, in_dim_pose=63, in_dim_trans=66, hidden=128):
        super().__init__()
        # POSE HEAD - 32,831 params; predicts root-relative residual (21x3)
        self.pose_trunk = nn.Sequential(
            nn.Linear(in_dim_pose, hidden), nn.GELU(),
            nn.Linear(hidden, hidden), nn.GELU(),
        )
        self.pose_head = nn.Linear(hidden, 63)
        nn.init.zeros_(self.pose_head.weight)
        nn.init.zeros_(self.pose_head.bias)
        # TRANSLATION HEAD - 25,475 params; wrist xyz correction
        # input = root-relative pose features (63) + raw root xyz (3) = 66 dim
        self.trans_trunk = nn.Sequential(
            nn.Linear(in_dim_trans, hidden), nn.GELU(),
            nn.Linear(hidden, hidden), nn.GELU(),
        )
        self.trans_head = nn.Linear(hidden, 3)
        nn.init.zeros_(self.trans_head.weight)
        nn.init.zeros_(self.trans_head.bias)

    def forward(self, pred):
        # pred: (B, 21, 3) - backbone joints in camera frame
        pred_root = pred[:, 0]                              # (B, 3)
        pose_rel  = (pred - pred_root.unsqueeze(1)).flatten(1)  # (B, 63)
        delta_pose  = self.pose_head(
            self.pose_trunk(pose_rel)).view(-1, 21, 3)      # (B, 21, 3)
        delta_trans = self.trans_head(self.trans_trunk(
            torch.cat([pose_rel, pred_root], -1)))          # (B, 3)
        return pred + delta_pose + delta_trans.unsqueeze(1) # broadcast trans
\end{verbatim}
\end{small}

\subsection{Per-layer parameter breakdown}
\label{appx:jrn-params}

\begin{table}[h]
\centering
\footnotesize
\begin{tabular}{lrr}
\toprule
Layer & Shape & Params \\
\midrule
Pose trunk Linear 1 & $63 \to 128$ & 8{,}192 \\
Pose trunk Linear 2 & $128 \to 128$ & 16{,}512 \\
Pose head (zero-init) & $128 \to 63$ & 8{,}127 \\
\midrule
\textbf{Pose subtotal} & & \textbf{32{,}831} \\
\midrule
Trans trunk Linear 1 & $66 \to 128$ & 8{,}576 \\
Trans trunk Linear 2 & $128 \to 128$ & 16{,}512 \\
Trans head (zero-init) & $128 \to 3$ & 387 \\
\midrule
\textbf{Trans subtotal} & & \textbf{25{,}475} \\
\midrule
\textbf{Total JRN} & & \textbf{58{,}306} \\
\bottomrule
\end{tabular}
\caption{\textbf{Per-layer parameter count of the JRN.} Linear-layer params $= d_{\text{in}} \cdot d_{\text{out}} + d_{\text{out}}$ (weight + bias). Total exactly matches the runtime-printed 58{,}306.}
\label{tab:jrn-params}
\end{table}

\subsection{Zero-initialization guarantee}
\label{appx:zero-init}

Both \texttt{pose\_head} and \texttt{trans\_head} are zero-initialized for both weight matrix \emph{and} bias. At training start, $\delta_{\text{pose}} = \mathbf{0}$ and $\delta_{\text{trans}} = \mathbf{0}$, so $\mathrm{JRN}(\text{pred}) \equiv \text{pred}$ identically. Two consequences:
\begin{itemize}
\item JRN matches the zero-shot baseline at initialization; training only moves JRN \emph{away} from the identity if gradients are informative. Empirically, after 60 epochs JRN improves all five metrics on all eight methods (Tab.~\ref{tab:headline}) without regressions.
\item Cross-machine or cross-environment determinism checks are unaffected by initialization: the same seed (42) plus the same data loader produces the same trained head, with no dependence on PRNG state at module-instantiation time.
\end{itemize}

\subsection{Why the two heads are orthogonal}
\label{appx:orthogonal}

We show that the pose head and translation head receive disjoint gradient signals from the five reported metrics, making the architecture decompositional rather than entangled.

Let $T(\delta) \in \mathbb{R}^{21 \times 3}$ denote rigid translation by $\delta \in \mathbb{R}^3$ broadcast to all 21 joints. The five metrics partition into two orthogonal classes under $T$.

\paragraph{Translation-invariant metrics.}
\begin{itemize}
    \item \textbf{PA-MPJPE} applies Umeyama Procrustes which removes translation before computing per-joint distance: $\partial / \partial \delta_{\text{trans}} = 0$.
    \item \textbf{MPJPE-RR} subtracts the root from both prediction and GT: $\partial / \partial \delta_{\text{trans}} = 0$.
    \item \textbf{PA-P2D} projects Procrustes-aligned 3D, so any rigid translation is absorbed during alignment: $\partial / \partial \delta_{\text{trans}} = 0$.
\end{itemize}
The translation head therefore receives \emph{exactly zero} gradient from these three metrics.

\paragraph{Translation-coupled metrics.}
\begin{itemize}
    \item \textbf{Root err} $= \| \hat{x}_0 - x_{\text{GT},0} \|$ depends only on the wrist correction. The pose head's training target $Y_{p}$ is identically zero at the wrist by construction ($Y_{p}[0] = (x_{\text{GT},0} - x_{\text{GT},0}) - (x_0 - x_0) = 0$), so $\delta_{\text{pose}}[0] \approx 0$ after training and Root err is left approximately unchanged by the pose head alone (e.g., WildHands JRN-pose-only Root err 550.4\,mm $\equiv$ ZS, per ablation).
    \item \textbf{MPJPE-abs} depends on \emph{both} heads: a linear translation through $T$, and the non-zero non-wrist pose corrections $\delta_{\text{pose}}[i{>}0]$ that move skeleton joints in absolute coordinates. Empirically, applying pose-only JRN can shift MPJPE-abs by ${\sim}20$\,mm even when Root err is unchanged (e.g., WildHands JRN-pose-only: 553.6\,mm vs ZS 533.3\,mm). The full JRN, with both heads trained, drives MPJPE-abs down 69-86\% across the six MANO-based backbones (Tab.~\ref{tab:headline}).
\end{itemize}

\paragraph{Empirical verification.} We compared two trained instances per backbone: (i)~JRN with the translation head frozen at zero (``JRN-pose only,'' equivalent to a 33K-parameter pose-only variant), and (ii)~the full 58K JRN. PA-MPJPE values were bit-identical across all six backbones (Table~\ref{tab:bit-identity}); the same holds for MPJPE-RR and PA-P2D (verified offline). This confirms that adding the translation head does not affect the translation-invariant metrics and does not leak supervision into the pose path.

\begin{table}[h]
\centering
\footnotesize
\begin{tabular}{lrrr}
\toprule
Backbone & JRN-pose only PA-MPJPE & JRN full PA-MPJPE & $\lvert$diff$\rvert$ \\
\midrule
WildHands & 10.4006 & 10.4006 & 0.0000 \\
JointTransformer & 10.6992 & 10.6992 & 0.0000 \\
HaMeR & 16.1102 & 16.1102 & 0.0000 \\
MeshGraphormer & 15.1184 & 15.1184 & 0.0000 \\
WiLoR & 15.9121 & 15.9121 & 0.0000 \\
SimpleHand & 21.1039 & 21.1039 & 0.0000 \\
\bottomrule
\end{tabular}
\caption{\textbf{PA-MPJPE bit-identity between pose-only JRN (33K) and full JRN (58K).} The 0.0000\,mm difference across all six backbones confirms that the translation head is orthogonal to the pose path under Procrustes alignment, and that the supervision targets do not leak into the pose head.}
\label{tab:bit-identity}
\end{table}

\section{JRN architecture diagram}
\label{appx:jrn-diagram}

The figure below illustrates the data flow through JRN. The top row shows the frozen-backbone pipeline that yields the $21 \times 3$ camera-frame prediction $x = f(I)$. Below it, $x$ feeds two parallel branches that are re-merged at a single sum node together with a skip from $x$ itself, giving the corrected output $\hat{x} = x + \delta_{p} + \delta_{t}\cdot\mathbf{1}_{21}$. The training loss is reproduced inside the figure for self-containment.

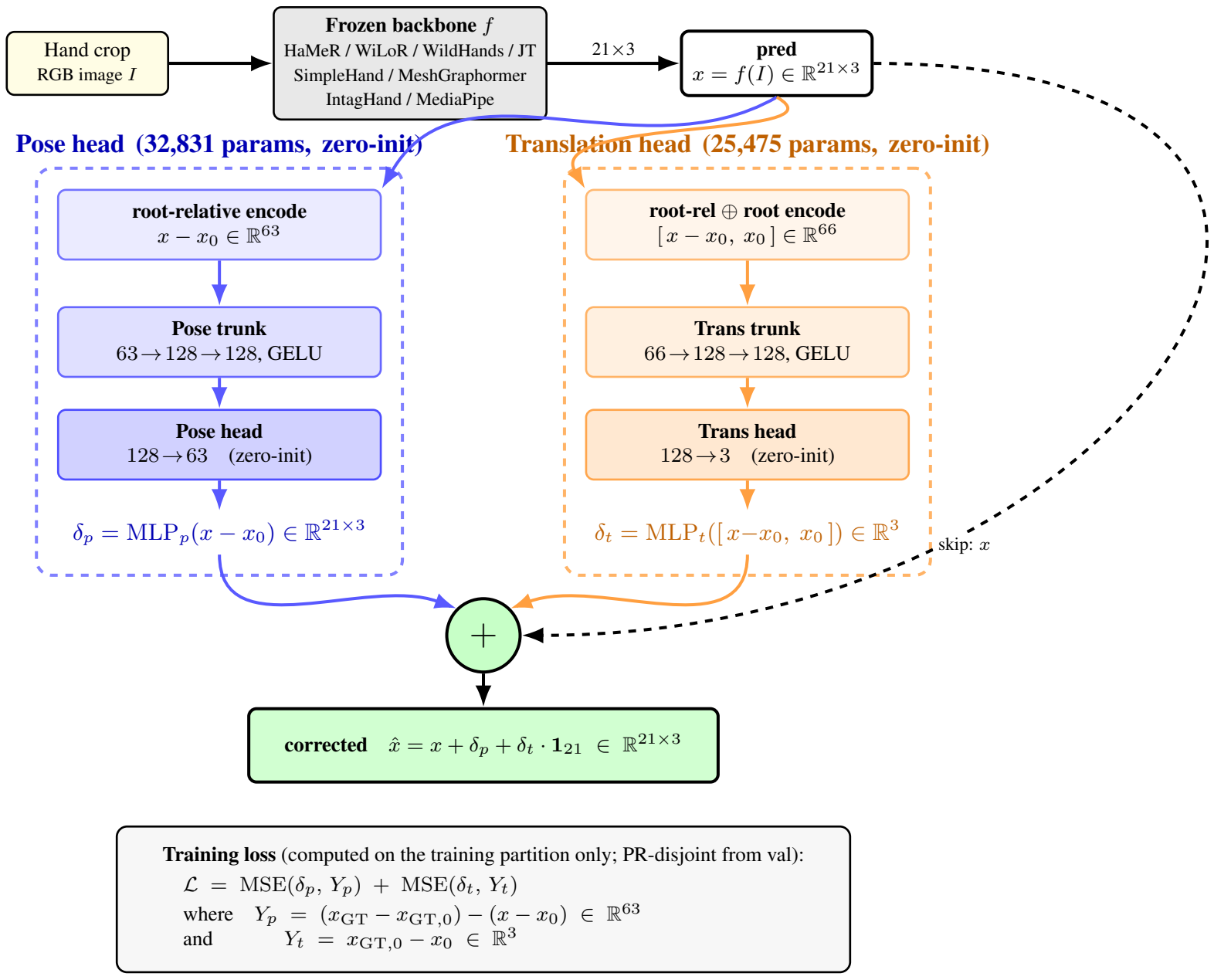
\begin{figure}[h]
\centering
\resizebox{0.92\linewidth}{!}{%
\begin{tikzpicture}[
    >=Latex,
    every node/.style={font=\footnotesize},
    box/.style={draw, thick, rounded corners=3pt, minimum height=0.85cm, align=center, inner sep=4pt},
    img/.style={box, fill=yellow!12, minimum width=2.2cm},
    bb/.style={box, fill=gray!18, minimum width=3.4cm},
    pred/.style={box, fill=white, very thick, minimum width=2.6cm},
    enc/.style={box, minimum width=4.4cm, minimum height=0.95cm},
    encP/.style={enc, fill=blue!8, draw=blue!55},
    encT/.style={enc, fill=orange!8, draw=orange!55},
    mlp/.style={box, minimum width=4.4cm, minimum height=0.95cm},
    mlpP/.style={mlp, fill=blue!12, draw=blue!60},
    mlpT/.style={mlp, fill=orange!12, draw=orange!60},
    head/.style={box, minimum width=4.4cm, minimum height=0.95cm},
    headP/.style={head, fill=blue!18, draw=blue!70},
    headT/.style={head, fill=orange!18, draw=orange!70},
    deltaP/.style={font=\small\bfseries, color=blue!70!black, align=center},
    deltaT/.style={font=\small\bfseries, color=orange!75!black, align=center},
    sumc/.style={circle, draw, very thick, minimum size=1.0cm, inner sep=0pt, fill=green!22, font=\Large\bfseries},
    outbox/.style={box, fill=green!18, very thick, minimum width=6.4cm, minimum height=1.0cm, font=\footnotesize\bfseries},
    arr/.style={->, very thick},
    arrP/.style={->, very thick, blue!65},
    arrT/.style={->, very thick, orange!75},
    edgelbl/.style={font=\scriptsize, inner sep=2pt, fill=white, rounded corners=1pt},
    losspanel/.style={draw, thick, rounded corners=4pt, fill=gray!6, inner sep=8pt, align=left, minimum width=10cm}
]

\node[img]   (img)  at (-5.4, 0) {Hand crop\\ \scriptsize RGB image $I$};
\node[bb]    (bb)   at (-1.0, 0) {\textbf{Frozen backbone $f$}\\ \scriptsize HaMeR / WiLoR / WildHands / JT\\ \scriptsize SimpleHand / MeshGraphormer\\ \scriptsize IntagHand / MediaPipe};
\node[pred]  (pred) at ( 4.0, 0) {\textbf{pred}\\ $x = f(I) \in \mathbb{R}^{21\times 3}$};
\draw[arr] (img) -- (bb);
\draw[arr] (bb) -- node[edgelbl, above] {21$\times$3} (pred);

\node[encP]  (rrel)  at (-3.6, -2.2) {\textbf{root-relative encode}\\ $x - x_0 \in \mathbb{R}^{63}$};
\node[encT]  (rmix)  at ( 3.6, -2.2) {\textbf{root-rel $\oplus$ root encode}\\ $[\,x - x_0,\ x_0\,] \in \mathbb{R}^{66}$};

\draw[arrP] (pred.south) to[out=-150, in=60] (rrel.north east);
\draw[arrT] (pred.south) to[out= -30, in=120] (rmix.north west);

\node[mlpP]  (poseT) at (-3.6, -3.8) {\textbf{Pose trunk}\\ $63 \!\to\! 128 \!\to\! 128$, GELU};
\node[headP] (poseH) at (-3.6, -5.2) {\textbf{Pose head}\\ $128 \!\to\! 63$\quad (zero-init)};
\draw[arrP] (rrel) -- (poseT);
\draw[arrP] (poseT) -- (poseH);
\node[deltaP] (deltaP) at (-3.6, -6.4) {$\delta_{p} = \mathrm{MLP}_{p}(x - x_{0}) \in \mathbb{R}^{21\times 3}$};
\draw[arrP] (poseH) -- (deltaP);

\node[mlpT]  (transT) at (3.6, -3.8) {\textbf{Trans trunk}\\ $66 \!\to\! 128 \!\to\! 128$, GELU};
\node[headT] (transH) at (3.6, -5.2) {\textbf{Trans head}\\ $128 \!\to\! 3$\quad (zero-init)};
\draw[arrT] (rmix) -- (transT);
\draw[arrT] (transT) -- (transH);
\node[deltaT] (deltaT) at (3.6, -6.4) {$\delta_{t} = \mathrm{MLP}_{t}([\,x{-}x_{0},\ x_{0}\,]) \in \mathbb{R}^{3}$};
\draw[arrT] (transH) -- (deltaT);

\begin{scope}[on background layer]
\node[fit=(rrel)(poseT)(poseH)(deltaP), draw=blue!50, dashed, very thick, rounded corners=6pt, inner sep=8pt] (poseGroup) {};
\node[fit=(rmix)(transT)(transH)(deltaT), draw=orange!60, dashed, very thick, rounded corners=6pt, inner sep=8pt] (transGroup) {};
\node[anchor=south, font=\bfseries\color{blue!70!black}] at (poseGroup.north) {Pose head \,(32{,}831 params,\, zero-init)};
\node[anchor=south, font=\bfseries\color{orange!75!black}] at (transGroup.north) {Translation head \,(25{,}475 params,\, zero-init)};
\end{scope}

\node[sumc] (sum) at (0, -7.8) {$+$};
\draw[arrP] (deltaP.south) to[out=-90, in=160] (sum.north west);
\draw[arrT] (deltaT.south) to[out=-90, in= 20] (sum.north east);
\draw[arr, dashed] (pred.east) to[out=0, in=0, looseness=2.4] node[edgelbl, right, near end] {skip: $x$} (sum.east);

\node[outbox] (out) at (0, -9.3) {corrected\quad $\hat{x} = x + \delta_{p} + \delta_{t}\cdot\mathbf{1}_{21} \;\in\; \mathbb{R}^{21\times 3}$};
\draw[arr, very thick] (sum) -- (out);

\node[losspanel] (losspanel) at (0, -11.4) {%
\textbf{Training loss} (computed on the training partition only; PR-disjoint from val):\\[2pt]
\quad $\mathcal{L} \;=\; \mathrm{MSE}(\delta_{p},\, Y_{p}) \;+\; \mathrm{MSE}(\delta_{t},\, Y_{t})$\\[2pt]
\quad where\quad $Y_{p} \;=\; (x_{\mathrm{GT}} - x_{\mathrm{GT},0}) - (x - x_{0}) \;\in\; \mathbb{R}^{63}$\\
\quad and\hphantom{where}\quad $Y_{t} \;=\; x_{\mathrm{GT},0} - x_{0} \;\in\; \mathbb{R}^{3}$
};

\end{tikzpicture}%
}
\caption{\textbf{JRN architecture.} The frozen backbone $f$ (any of the eight evaluated methods) maps the input image $I$ to a 21-joint prediction $x \in \mathbb{R}^{21\times 3}$ in camera frame. JRN is a 58{,}306-parameter two-headed MLP residual corrector. The pose head (blue, 32{,}831 params) consumes the root-relative pose $x - x_{0} \in \mathbb{R}^{63}$ and emits a $21\times 3$ root-relative residual $\delta_{p}$. The translation head (orange, 25{,}475 params) consumes the root-relative pose concatenated with the raw wrist xyz, $[\,x - x_{0},\, x_{0}\,] \in \mathbb{R}^{66}$, and emits a 3-D wrist correction $\delta_{t}$ broadcast to all 21 joints via $\mathbf{1}_{21}$. The corrected output is $\hat{x} = x + \delta_{p} + \delta_{t}\cdot\mathbf{1}_{21}$. Both heads are zero-initialized, so JRN equals the identity at $t{=}0$. The training loss panel reproduces the formal definition; targets are computed from the training partition only. Pose head and translation head are orthogonal on four of the five reported metrics (Appx.~\ref{appx:orthogonal}): the pose head drives PA-MPJPE / MPJPE-RR / PA-P2D (translation-invariant under Procrustes), while the translation head drives Root err. MPJPE-abs depends on both heads; the full JRN trains them jointly.}
\label{fig:jrn-arch}
\end{figure}

\section{Training hyperparameters}
\label{appx:training}

JRN training settings are identical across all eight methods. We use the same script and configuration for every backbone; only the input zero-shot prediction file differs. The script asserts $\text{train\_PRs} \cap \text{val\_PRs} = \emptyset$ at runtime before any forward pass, so a leak from the validation partition would crash the script rather than silently inflate metrics.

\begin{table}[h]
\centering
\footnotesize
\begin{tabular}{ll}
\toprule
Hyperparameter & Value \\
\midrule
Optimizer & AdamW \\
Learning rate & $1 \times 10^{-3}$ \\
Weight decay & $1 \times 10^{-4}$ \\
Batch size & 512 \\
Epochs & 60 \\
LR schedule & constant \\
Random seed & 42 (\texttt{torch.manual\_seed}, \texttt{np.random.seed}) \\
Pose target $Y_p$ & $(x_{\text{GT}} - x_{\text{GT},0}) - (x - x_0)$ \\
Translation target $Y_t$ & $x_{\text{GT},0} - x_0$ \\
Loss & $\mathrm{MSE}(\delta_p, Y_p) + \mathrm{MSE}(\delta_t, Y_t)$ (equal weights) \\
Early stopping & none \\
Validation-based selection & none \\
Hyperparameter search & none \\
Training device & 1$\times$ NVIDIA A100 80\,GB \\
Wallclock per backbone & ${\approx}8$\,s for 60 epochs \\
Train-data memory footprint & $41{,}909 \times 66 \times 4$\,B $\approx 11$\,MB \\
Trainable parameters & 58{,}306 (Tab.~\ref{tab:jrn-params}) \\
\bottomrule
\end{tabular}
\caption{\textbf{JRN training configuration.} Identical for all eight methods (HaMeR, WiLoR, WildHands, JointTransformer, MeshGraphormer, SimpleHand, IntagHand, MediaPipe Hands). The seed is set inside the training script and overrides any external state, so external orchestration cannot perturb results.}
\label{tab:hyperparams}
\end{table}

\paragraph{Determinism check.} We re-ran WildHands training with an outer seed of 43 (set in the calling shell) while leaving the in-script seed at 42; results were bit-identical to the plain seed-42 run (PA-MPJPE 10.4006, MPJPE-abs 75.79 in both cases). This confirms the script's internal seeding dominates and that reproduction is robust to external orchestration.

\section{Backbone provenance and preprocessing}
\label{appx:backbones}

All eight methods are \emph{frozen} during JRN training: weights are loaded once and never updated. Checkpoints are the original authors' public releases. None of the eight checkpoints were fine-tuned on EgoSurgHands.

\paragraph{Pretrained backbones.} \textbf{HaMeR}~\cite{hamer} (CVPR 2024) is a ViT-Huge MANO regressor pretrained on a mixture of FreiHAND, HO-3D, MTC, RHD, InterHand2.6M, H2O3D, DexYCB, and complementary 2D-pose corpora (COCO-WholeBody, Halpe, MPII NZSL); we treat it as our non-egocentric reference. \textbf{WiLoR}~\cite{wilor} (CVPR 2025) couples a real-time CNN hand detector, trained on a 2M-image in-the-wild dataset, with a transformer-based MANO regressor trained on a 4.2M-image multi-dataset mix. \textbf{WildHands}~\cite{wildhands} (ECCV 2024) is an egocentric hand-pose model that adds an intrinsics-aware positional encoding (KPE) to a hand-pose backbone, trained on egocentric ARCTIC and AssemblyHands images with auxiliary supervision from EPIC-Kitchens and Ego4D; we use it as the closest publicly available ego-trained model for the ego-vs-non-ego comparison against HaMeR. \textbf{JointTransformer}~\cite{jt} (HANDS'2023 ARCTIC Challenge winner @ ICCV) pairs a frozen DINOv2 ViT-G encoder with per-joint decoder queries (decoder depth 12); its decoder is trained on ARCTIC's allocentric task and then fine-tuned on the egocentric task, giving a second ego-trained backbone with a different architecture. \textbf{MeshGraphormer}~\cite{meshgraphormer} (ICCV 2021) combines graph and transformer reasoning on a third-person corpus, and \textbf{SimpleHand}~\cite{simplehand} (CVPR 2024) is a lightweight monocular regressor used as a minimum-capacity baseline. \textbf{IntagHand}~\cite{li2022intaghand} (CVPR 2022) reconstructs two interacting hands jointly from a single image with an interacting attention graph, addressing the bimanual character of surgical scenes. \textbf{MediaPipe Hands}~\cite{mediapipe} is a keypoint-only method that regresses 21 hand landmarks without a MANO mesh; it detects a hand on 24.1\% of evaluation rows, so its errors are computed on detected rows only. All checkpoints are obtained from the original authors' public releases; no Aria pad data appears in any backbone's pretraining mixture.

\paragraph{Preprocessing.} Every backbone receives the same fixed crop of the 1408$\times$1408 Aria frame (pixels 300 to 1200 in both axes) rather than a per-frame hand detection; the dataset's per-hand bounding boxes are not used by the evaluation. The eval-side wrappers we publish (one per backbone) handle (i)~image normalization with ImageNet statistics, (ii)~resize to the model's expected resolution (224$\times$224 for ViT-G; 256$\times$256 for ViT-Huge variants), and (iii)~post-prediction MANO-to-OpenPose joint reordering where applicable. We applied a single bug-fix to the WiLoR wrapper to remove a double-application of \texttt{MANO\_TO\_OPENPOSE}; all other wrappers reproduce upstream behavior unchanged, and the released WiLoR wrapper includes the fix.

\paragraph{Output convention.} All wrappers return predictions in the OpenPose 21-joint convention, $\big[$wrist, thumb-CMC, thumb-MCP, thumb-IP, thumb-tip, index-MCP, index-PIP, index-DIP, index-tip, middle-MCP, $\dots$, pinky-tip$\big]$. The eval pipeline drops index 1 (thumb-CMC) at metric time because there is no Aria MPS analog (\S\ref{sec:metrics}); the released per-frame predictions retain all 21 joints so downstream users can reproduce alternative joint subsets if desired.

\section{Annotation tool}
\label{appx:annotator}

The human-validated evaluation split (\S\ref{sec:iaa}) was produced with a custom keypoint-verification web application that we release as open source alongside the dataset. The tool is dataset-agnostic by design: any image set with 2D-keypoint overlays can be ingested via a single JSON manifest, so it generalizes beyond surgical hand pose to any landmark- or skeleton-verification task. This appendix documents the user-facing functionality; implementation-stack details are summarized at the end.

\begin{figure}[h]
  \centering
  \includegraphics[width=\linewidth]{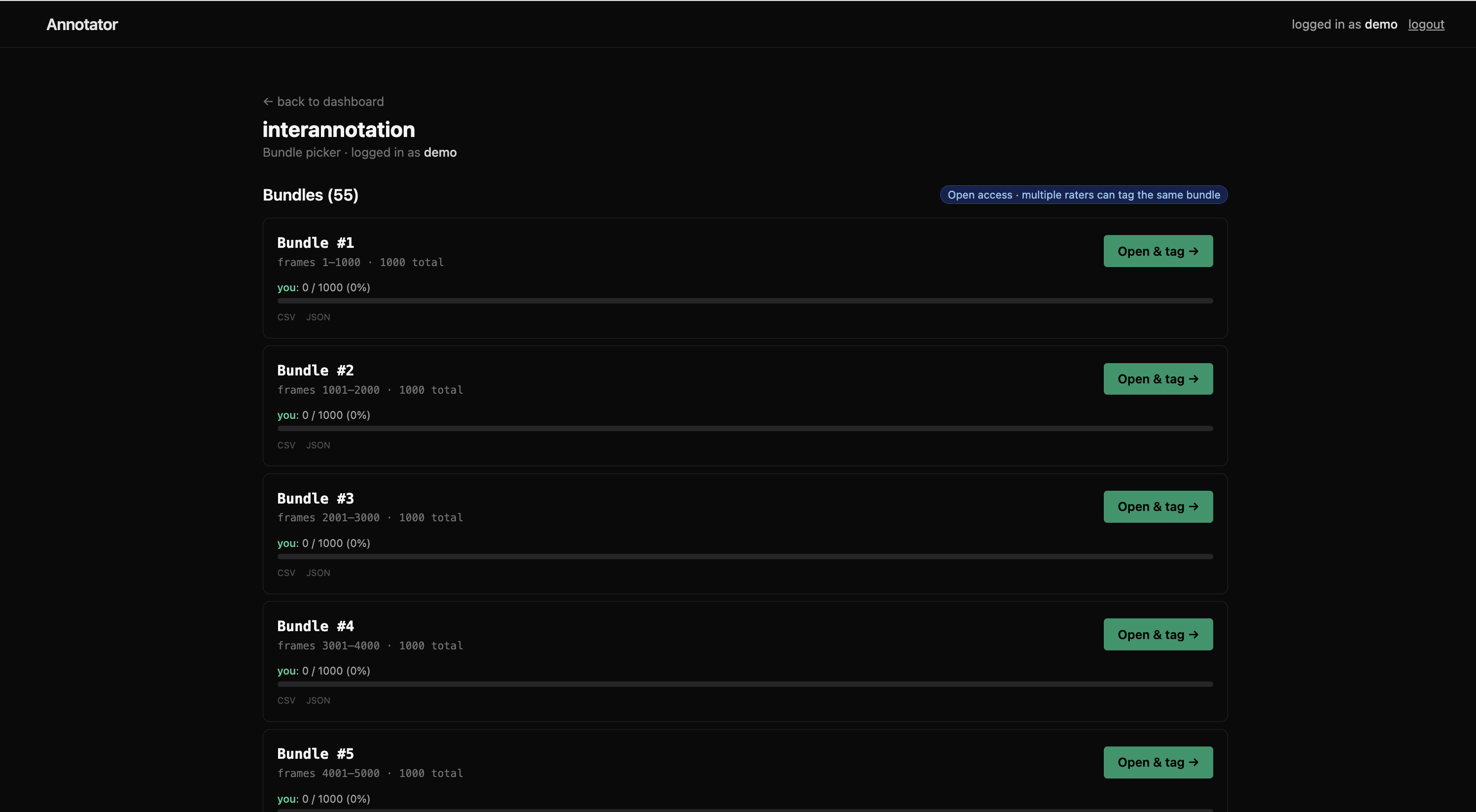}
  \caption{\textbf{Annotation tool, bundle picker.} Each evaluation pool is split into fixed-size bundles (1{,}000 frames each); annotators select a bundle to work on and resume across sessions. Per-bundle progress is persisted server-side. The ``open access'' flag enables multiple raters to independently tag the same bundle, which is the substrate for the multi-rater review and inter-annotator-agreement audit described in \S\ref{sec:iaa}.}
  \label{fig:annotator-dashboard}
\end{figure}

\paragraph{Bundle dispatcher (Fig.~\ref{fig:annotator-dashboard}).} On login, each annotator lands on a per-dataset \emph{bundle picker}: the candidate evaluation pool is split into fixed-size bundles (we use 1{,}000 frames per bundle), and an open-access flag allows multiple raters to independently tag the same bundle for inter-rater agreement. Per-bundle progress (\texttt{your-done / total}) is persisted across sessions; CSV and JSONL exports are exposed inline on each bundle row. For EgoSurgHands we used $55$ bundles for the full inter-annotation pool.

\begin{figure}[h]
  \centering
  \includegraphics[width=\linewidth]{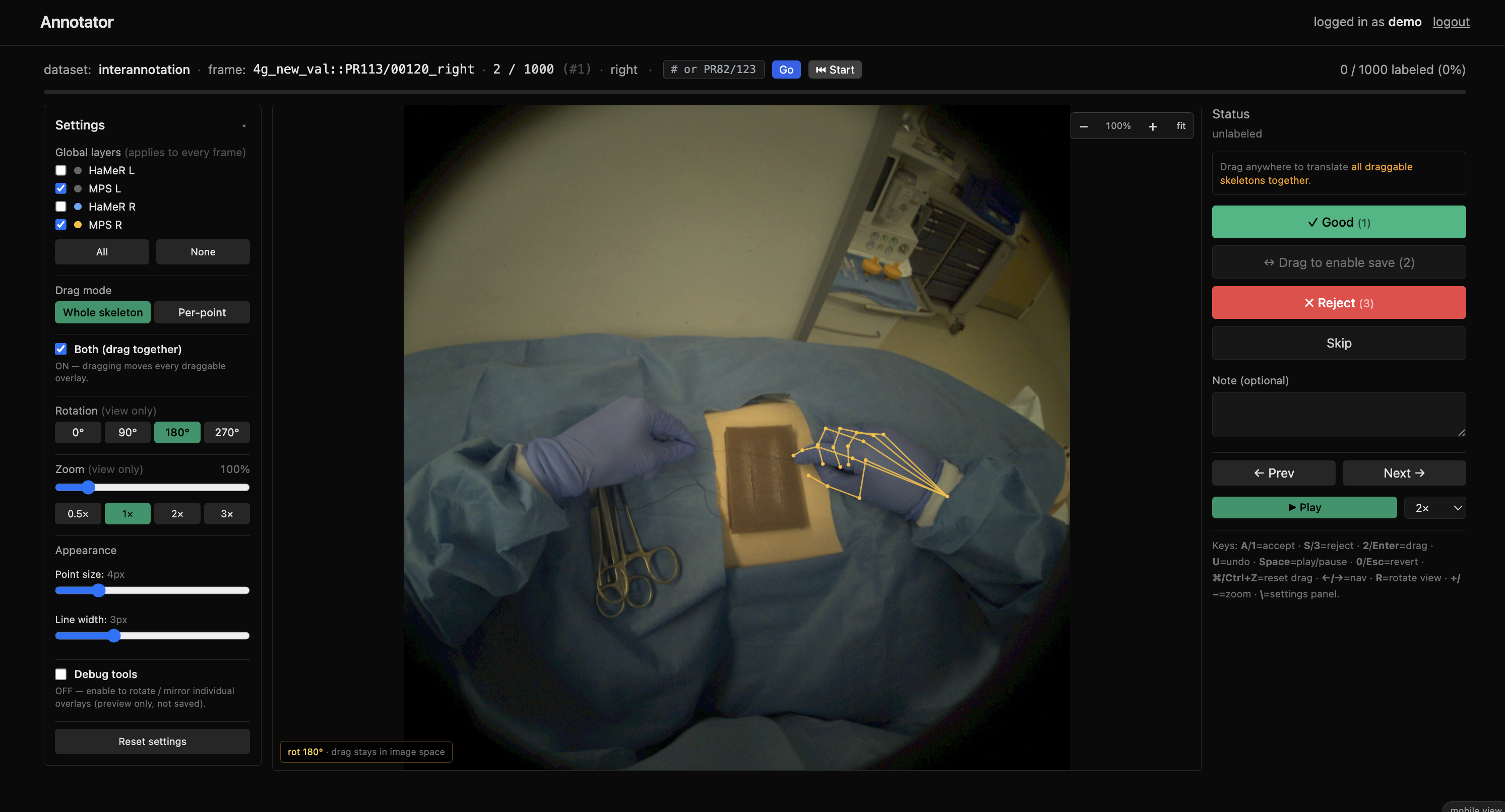}
  \caption{\textbf{Annotation tool, frame canvas.} Each frame is rendered with all available skeleton overlays (HaMeR-left/right, MPS-left/right) toggleable individually from the Settings panel (left). Reviewers translate skeletons by direct drag (whole-skeleton or per-point modes; only the whole-skeleton translation is stored), rotate the canvas in 90\textdegree~increments, and zoom from $0.5\times$ to $3\times$. The Status panel (right) holds the three-state labeling buttons (Good / Drag-and-save / Reject) plus Skip, an optional free-text note, prev/next navigation, and a keyboard-shortcut reference. The example frame shows a right-hand suturing pose with the MPS-right overlay (orange) about to be drag-corrected by the rater.}
  \label{fig:annotator-canvas}
\end{figure}

\paragraph{Annotation canvas (Fig.~\ref{fig:annotator-canvas}).} Each frame is rendered with all available overlays (HaMeR-left, HaMeR-right, MPS-left, MPS-right) drawn on top of the rectified RGB crop. Reviewers can:
\begin{itemize}
\item \textbf{Toggle overlays} individually (e.g., view only MPS-right, hide HaMeR overlays).
\item \textbf{Translate skeletons} by direct drag, in either ``whole-skeleton'' or ``per-point'' mode; a paired \emph{drag together} option moves both hands simultaneously to handle systematic offsets in MPS projections. The drag offset $(\Delta x, \Delta y)$ is recorded in image-pixel units; only the whole-skeleton translation is stored, so a recorded correction is a global image-plane translation (\S\ref{sec:iaa}).
\item \textbf{Rotate} the canvas to any of \{0, 90, 180, 270\}\textdegree~for visual clarity (drag stays in image-space coordinates regardless of view rotation).
\item \textbf{Zoom} from $0.5\times$ to $3\times$ for fine-grained joint-level inspection.
\item Adjust \textbf{point size} and \textbf{line width} of the rendered skeleton, plus enable a debug overlay that exposes per-overlay rotation and mirroring controls (preview only, not saved).
\end{itemize}

\paragraph{Three-state labeling.} Each frame receives one of three primary states, plus a \emph{Skip} option for low-confidence frames returned to later: (i)~\textbf{Good}: the overlay aligns with the real hand as is; (ii)~\textbf{Drag-and-save}: the skeleton shape is correct but its position is offset, requiring a translation to align it (the drag offset is saved); (iii)~\textbf{Reject}: the skeleton shape itself is wrong, indicating a tracker failure that should be excluded from evaluation. A free-text per-frame note can be attached to capture the reason for rejection (e.g., motion blur, partial occlusion, hand-side mislabel).

\paragraph{Keyboard-driven workflow.} The full review loop is driveable from the keyboard, which approximately doubles annotation throughput in practice: \texttt{A}/\texttt{1} = accept, \texttt{S}/\texttt{3} = reject, \texttt{2}/\texttt{Enter} = enter or save drag, \texttt{U} = undo, \texttt{Space} = play/pause autoplay, \texttt{0}/\texttt{Esc} = revert drag, \texttt{Cmd/Ctrl+Z} = reset drag, \texttt{$\leftarrow$/$\rightarrow$} = navigate frames, \texttt{R} = rotate view, \texttt{+/$-$} = zoom, \texttt{\textbackslash} = toggle settings panel.

\paragraph{Multi-rater support.} The data model stores one annotation per (frame, user) pair, with a unique \texttt{(frameId, userId)} constraint, so multiple raters can independently label the same frame without overwriting each other. This is what enables the multi-rater workflow described in \S\ref{sec:iaa} (primary review, examiner cross-review, surgeon adjudication, plus the 500-frame stratified audit subset for inter-annotator agreement). Per-user dashboards show resumable progress, so reviewers can stop and resume mid-bundle.

\paragraph{Export and downstream use.} Two export formats are emitted per bundle: (i)~CSV, one row per (frame, user) annotation with state, drag offset, note, and timestamp; (ii)~JSONL, the same record plus the full overlay payload for downstream re-rendering. The exported records are the input to the aggregation pipeline that constructs the final evaluation partition of 9{,}476 hand instances.

\paragraph{Implementation stack.} Next.js 14 (app router) + React 18 with TypeScript; Tailwind for styling; Prisma + SQLite for the data model (\texttt{User}, \texttt{Dataset}, \texttt{Frame}, \texttt{Annotation}); HMAC-signed session cookies, no external service dependencies. The whole tool runs on a single machine with one \texttt{npm run dev} command; large-scale deployments substitute the SQLite backing store with PostgreSQL via the Prisma provider switch. Source code is released alongside the dataset under the same license.

\subsection{Annotation audit details}
\label{app:iaa}

This subsection gives the full audit statistics summarized in \S\ref{sec:iaa}. Reject actions carry reason codes. Most frames received single-rater primary review (standard at this scale); we mitigate via examiner cross-review, surgeon adjudication, and a stratified double-annotation subset for inter-rater agreement.

\paragraph{Correction and rejection rates.} Across 9{,}476 review decisions, 43.1\% of frames were accepted as-is, 51.9\% received a drag correction, and 5.1\% were rejected. Across the 4{,}914 corrections, the median correction magnitude is 8.2\,mm after conversion using per-frame depth and Aria intrinsics; the mean is 9.8\,mm, and the principal annotators' medians range from 7.8 to 10.3\,mm. Frames judged correct carry a zero offset by definition; including them, the mean correction across all 9{,}476 reviewed hand instances is 5.3\,mm. An offset is recorded only after an annotator judges an overlay to require correction, so 8.2\,mm is the median correction conditional on flagging rather than the average error over the full evaluation population, and the stored operation is a global image-plane translation converted to metric units, rather than an independently annotated per-joint 3D displacement. The 51.9\% figure therefore reflects how often annotators made a sub-centimeter adjustment rather than how often a label failed; the rejection rate is 5.1\%. We use this analysis to quantify the magnitude of human intervention, not as a mocap-equivalent estimate of per-joint 3D error.

\paragraph{Inter-annotator agreement.} Because naturally double-annotated cases were sparse, we constructed a 500-frame stratified audit subset drawn from the evaluation split, in which three annotators independently made a keep/reject decision on every frame. Pairwise agreement is 70.0\%, 71.4\%, and 77.1\%; pooled pairwise agreement is 72.9\%; pooled Cohen's $\kappa$ is 0.423, PABAK 0.457, Gwet AC1 0.489, Fleiss $\kappa$ 0.421, and Krippendorff $\alpha$ 0.422. Surgeon adjudication changed the final label on 55.5\% of escalated frames. Escalation selects for the frames where primary review was uncertain, so a reversal rate near half on that subset reflects the difficulty of the cases chosen for escalation rather than the state of the evaluation split as a whole; the escalated set is small relative to the partition.

\section{Datasheet for EgoSurgHands}
\label{app:datasheet}

The full Datasheet for Datasets~\cite{datasheets} (Gebru et al.\ template) ships with the release at \url{https://huggingface.co/datasets/surgicalinformatics/EgoSurgHands} as \texttt{DATASHEET.md}. We summarize its seven sections here.

\paragraph{Motivation.} EgoSurgHands was created to enable benchmarking of monocular egocentric 3D hand-pose estimation in surgical training environments, a domain poorly covered by existing egocentric hand corpora (cooking, sports, factory assembly). Operationally relevant downstream applications include surgical-skill assessment, AR-guided training, and surgical robot learning from demonstration. Created by the authors (Harvard Medical School and Beth Israel Deaconess Medical Center); GPU compute was provided by Nebius and through funding from the Josiah Macy Jr.\ Foundation Artificial Intelligence in Medical Education Grant, and the Aria headsets were an in-kind donation from Meta Platforms, Inc.

\paragraph{Composition.} Each instance is one (frame, hand) pair: an Aria 1408$\times$1408 ego RGB frame, 21-joint 3D hand pose in camera frame (Aria MPS sensor-fused pseudo-ground truth; not mocap-grade), 2D projection of joints, hand-side label, joint-hull bounding box (with 16.5\% padding), and per-frame metadata (wearer, task, glove, experience tier, recording day). Total: 51{,}385 hand-instance rows (41{,}909 train + 9{,}476 human-validated evaluation), 32{,}646 unique frames after L/R pairing. No PII; all wearers are consenting adult medical trainees.

\paragraph{Collection process.} Direct observation. Surgeons-in-training wore Project Aria research-edition glasses (1408$\times$1408 fisheye RGB at 30\,fps) while performing 13 standard suturing tasks on synthetic-skin training pads. Aria MPS, Meta's cloud post-processing service, fused the device's sensors (visual-inertial SLAM, multi-camera stereo, online calibration) into 21-joint 3D hand poses. Captured across days at medical-training simulation center (Beth Israel Deaconess Medical Center, Boston). No patient data, third-party PII, faces, or clinical-systems content is captured.

\paragraph{Preprocessing / cleaning / labeling.} Seven-step pipeline: (1) raw 30\,fps Aria recordings; (2) MPS sensor fusion (cloud post-processing) to 21-joint 3D pose; (3) \texttt{has\_eval\_gt} valid-projection filter (1{,}047 train rows dropped, 0 val rows dropped); (4) bbox = joint-hull + 16.5\% padding; (5) \texttt{recording\_day} remapping (real ISO dates removed); (6) multi-rater human review (primary review, examiner review, surgeon adjudication of flagged disagreements) on the validation split; (7) \texttt{has\_eval\_gt} column dropped post-filter. Review details and the annotation audit are in \S\ref{sec:iaa}.

\paragraph{Uses.} \textbf{Validated:} zero-shot benchmarking of monocular hand-pose models on egocentric surgical training video; within-dataset MPS-label adaptation baselines such as JRN (this paper); in-distribution per-task / per-wearer / per-glove analysis (glove color and wearer are confounded for P4/brown). \textbf{Not validated, prohibited:} performance scoring for hiring or licensure decisions, live-OR deployment without further domain-specific fine-tuning, person re-identification, cross-site or cross-population generalization claims, and fine-grained ranking of corrected models whose differences fall below the roughly 1\,cm label resolution scale.

\paragraph{Distribution.} Public release on Hugging Face under CC BY-NC-SA 4.0 (non-commercial, share-alike, attribution). Upstream Project Aria research-license terms also apply (\url{https://www.projectaria.com/datasets/dataset-license-agreement/}). Hosted at \url{https://huggingface.co/datasets/surgicalinformatics/EgoSurgHands} for at least 24 months from publication. Persistent identifier (HF DOI) attached on first stable release.

\paragraph{Maintenance.} Maintainer contact via the HF repository's Discussions tab. Versioning follows semantic conventions; v1 freezes the rows reported in this paper. Future releases (e.g., cross-site extensions) will use semver bumps with explicit changelog. Errata are tracked on the HF repository.

\section{Croissant metadata}
\label{app:croissant}

A complete Croissant~\cite{croissant} metadata file conforming to the MLCommons Croissant 1.1 specification ships with the release as \texttt{croissant.json} at \url{https://huggingface.co/datasets/surgicalinformatics/EgoSurgHands}. The file declares: \texttt{@context} (Schema.org + MLCommons Croissant + DataCite Terminology URIs); \texttt{@type} \texttt{sc:Dataset}; \texttt{name} = \texttt{EgoSurgHands}; \texttt{version} = \texttt{1.1.0}; \texttt{license} = \texttt{CC BY-NC-SA 4.0}; \texttt{url} pointing to the HF dataset page; \texttt{conformsTo} = \texttt{http://mlcommons.org/croissant/1.1}; creator (the authors; Harvard Medical School and Beth Israel Deaconess Medical Center); \texttt{isAccessibleForFree} = \texttt{true}; \texttt{isSyntheticData} = \texttt{false}.

\paragraph{Distribution.} Two \texttt{cr:FileSet} entries: \texttt{train-parquets} (12 Apache Parquet shards, 41{,}909 rows) and \texttt{validation-parquets} (3 Apache Parquet shards, 9{,}476 rows). Both with \texttt{encodingFormat} = \texttt{application/x-parquet}.

\paragraph{RecordSets.} Two \texttt{cr:RecordSet} entries (\texttt{train} and \texttt{validation}, 41 fields each), each declaring per-column \texttt{name}, \texttt{description}, \texttt{dataType}, and source-field references. Columns include \texttt{frame\_key}, \texttt{image} (sc:ImageObject), \texttt{intrinsics} / \texttt{extrinsics} / \texttt{bbox} / \texttt{joints\_3d} / \texttt{joints\_2d} (sc:Float repeated), \texttt{hand\_side} / \texttt{pr} / \texttt{wearer\_id} / \texttt{experience} / \texttt{glove\_color} / \texttt{task} / \texttt{recording\_day} (sc:Text), and \texttt{frame\_id} / \texttt{frame\_ts\_us} (sc:Integer). Version 1.1.0 adds per-hand MPS confidence, image-time and fisheye-projected labels, per-frame calibration, SLAM pose, and eye-gaze columns. The Responsible-AI fields are populated; we transcribe them in Appx.~\ref{app:rai}.

\section{Responsible AI considerations}
\label{app:rai}

We report the eight Responsible-AI fields required by the NeurIPS 2026 Evaluations \& Datasets track, followed by three further Croissant-RAI fields; \texttt{croissant.json} in the release carries all of them.

\paragraph{rai:dataLimitations.} Single-site capture; results may not generalize across institutions. Four wearers; under-represents left-handed and attending demographics. Recordings on synthetic-skin training pads, not live-OR data. Egocentric-only viewpoint. Monocular RGB. Evaluation labels are sensor-fused, human-validated pseudo-ground truth, and the original adaptation split is within-corpus.

\paragraph{rai:dataBiases.} All brown-glove evaluation hand instances ($n{=}661$) come from wearer P4, so glove color and wearer are confounded for this stratum. One left-handed wearer (P2) out of four. Single-site lighting and camera distribution.

\paragraph{rai:personalSensitiveInformation.} No patient data, third-party PII, faces, name tags, hospital-systems screens, or clinical-systems content is captured. Wearer IDs are de-identified to \texttt{P1}-\texttt{P4}; absolute timestamps are removed. The upstream Aria audio stream is not exported.

\paragraph{rai:dataUseCases.} Validated: zero-shot benchmarking of monocular hand-pose models on egocentric surgical training video; within-dataset MPS-label adaptation baselines; in-distribution per-task, per-wearer, and per-glove analysis. Not validated: performance scoring for hiring or licensure, live-OR deployment, person re-identification, cross-site or cross-population generalization claims, and fine-grained ranking of corrected models below the roughly 1\,cm label resolution scale (\S\ref{sec:labelquality}).

\paragraph{rai:dataSocialImpact.} \emph{Positive:} enables surgical-skill assessment, AR-guided trainee feedback, and surgical robot learning from demonstration. \emph{Negative:} potential surveillance of trainees; misuse for non-validated performance scoring; unsanctioned application of pad-trained models to live OR data. Mitigations: license restricts to non-commercial use; the datasheet flags scoring-for-hiring and live-OR deployment as prohibited.

\paragraph{rai:hasSyntheticData.} \texttt{false}. All frames are real Aria recordings of real adult trainees on synthetic-skin training pads. No synthetic-data augmentation in this release.

\paragraph{prov:wasDerivedFrom.} No source dataset: EgoSurgHands is an original capture. Frames and labels come from the authors' own Project Aria recordings and the corresponding Meta MPS outputs.

\paragraph{prov:wasGeneratedBy.} The seven-step pipeline in the datasheet's preprocessing paragraph (Appx.~\ref{app:datasheet}): MPS sensor fusion to 21-joint 3D pose, valid-projection filtering, bounding-box derivation, recording-date remapping, and multi-rater human review of the evaluation partition.

\paragraph{rai:dataCollection.} Aria MPS sensor fusion (Meta's cloud post-processing of the recordings) producing 21-joint 3D hand poses at 30\,Hz; followed by valid-projection quality filter and (for the validation split) multi-rater human review with surgeon adjudication of flagged disagreements. All recordings are on synthetic-skin training pads.

\paragraph{rai:dataAnnotationProtocol.} MPS auto-annotation followed by multi-rater human validation on the 9{,}476-row validation partition: primary review, examiner review, and surgeon adjudication of flagged disagreements. Inter-annotator agreement on a 500-frame stratified audit subset: Cohen's $\kappa$ 0.423, Fleiss $\kappa$ 0.421, Krippendorff $\alpha$ 0.422 (\S\ref{sec:iaa}).

\paragraph{rai:annotatorDemographics.} Annotators are drawn from the surgical and hand-pose research communities

\paragraph{IRB / ethics.} Because the dataset contains no patient data and only consenting adult participants performing routine surgical training exercises on synthetic-skin pads, formal IRB review was not required under the institution's non-human-subjects-research policy.

\section{Additional figures}
\label{app:figs}

\begin{figure}[h]
  \centering
  \includegraphics[width=\linewidth]{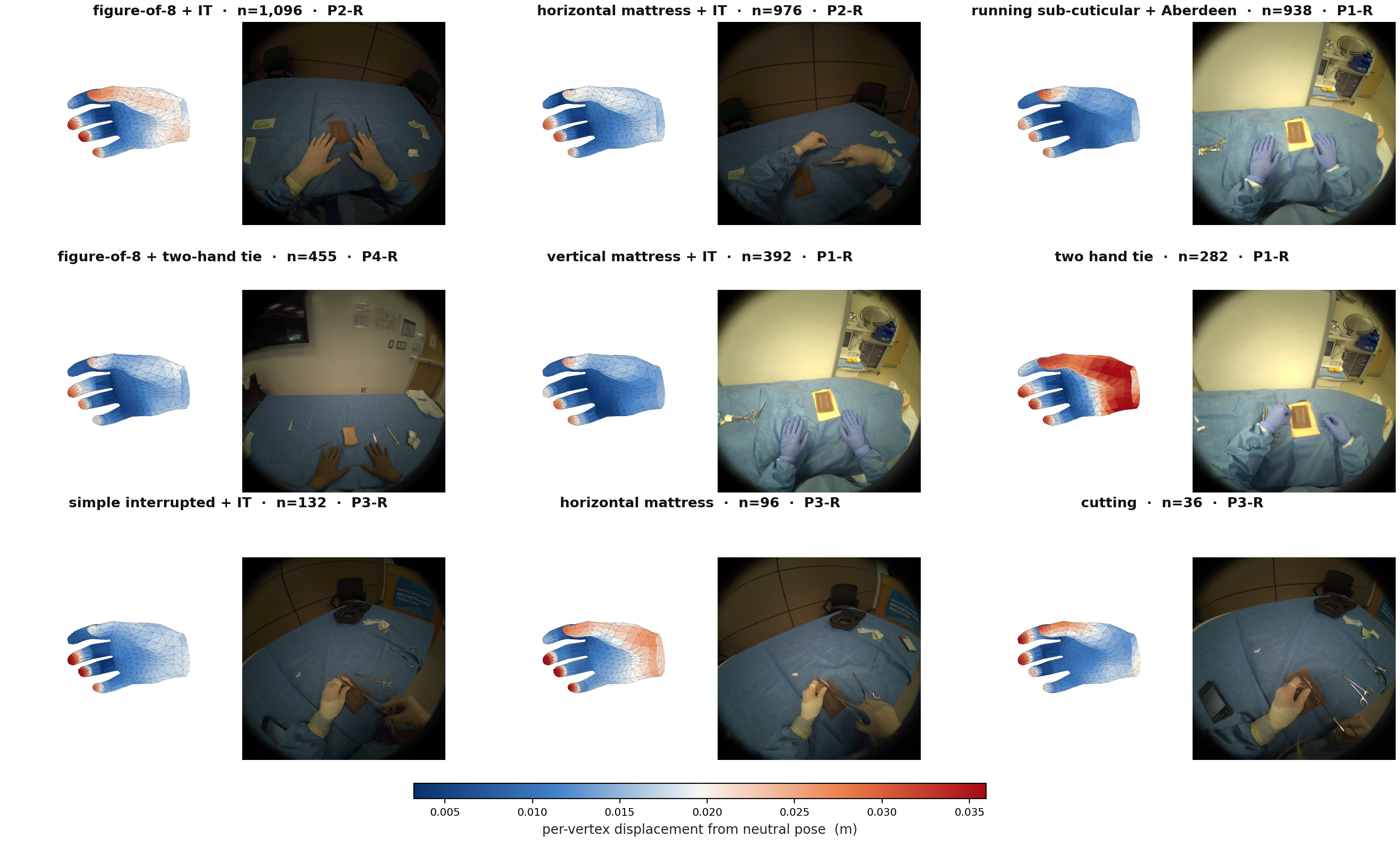}
  \caption{\textbf{Per-task hand-pose fingerprints.} For each surgical training task we show (left) the canonical neutral MANO mesh colored by per-vertex displacement from neutral, computed as the rigid-aligned distance between the per-task median MANO surface and the neutral T-pose, and (right) a representative raw ego frame. Red = high-displacement regions; blue = parts that stay near neutral. Colormap is shared globally so panels are directly comparable. Surgical tasks produce visually distinct surface fingerprints recovered from a single median pose.}
  \label{fig:fingerprints}
\end{figure}

\paragraph{Dataset statistics.}
\label{app:stats}
EgoSurgHands is balanced across the diversity axes that distinguish surgical training from generic egocentric data. The evaluation partition is jointly stratified by wearer (P1: 4{,}694, P2: 3{,}198, P3: 923, P4: 661 hand instances), experience tier (PGY3: 5{,}355, MS1: 3{,}198, PGY0: 923), glove color (blue 4{,}694, white 4{,}121, brown 661), and task family. Hand-side counts are approximately even within both splits. Per-PR row counts vary substantially across the evaluation set (min 75, max 2{,}162), and per-task and per-day distributions are reported in the supplementary material. A per-task qualitative summary of hand-pose fingerprints across the 13 surgical training tasks (median MANO surface colored by per-vertex displacement from the neutral hand mesh, with paired raw ego frames) is shown in Fig.~\ref{fig:fingerprints}; surgical tasks produce visually distinct surface fingerprints recovered from the dataset alone, showing that EgoSurgHands captures task-discriminative motion patterns at the hand-mesh level.

\begin{figure}[h]
  \centering
  \includegraphics[width=0.92\linewidth]{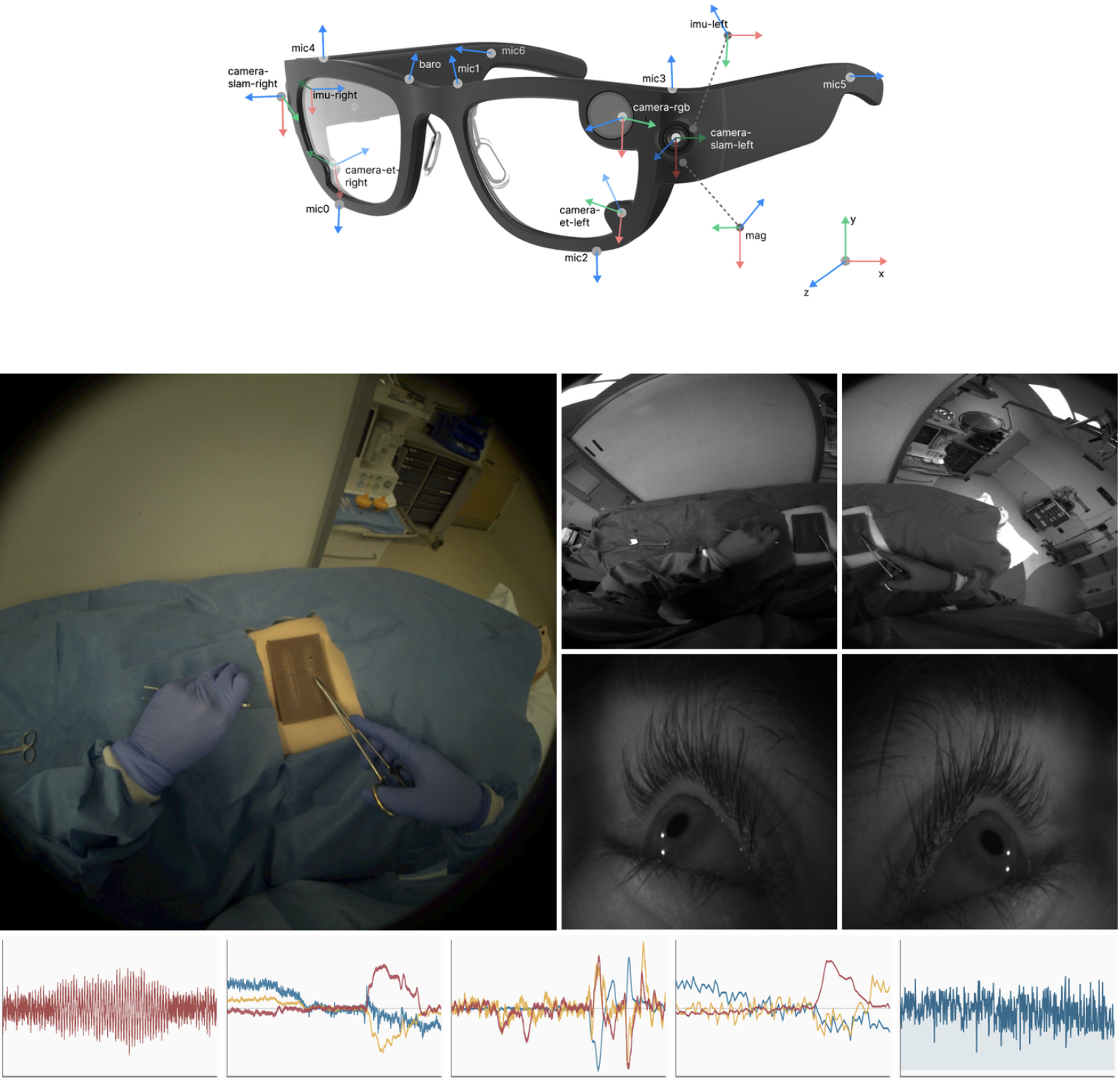}
  \caption{\textbf{Sensor streams recorded by the Project Aria device.}}
  \label{fig:aria-device}
\end{figure}

\begin{figure}[h]
  \centering
  \includegraphics[width=0.62\linewidth]{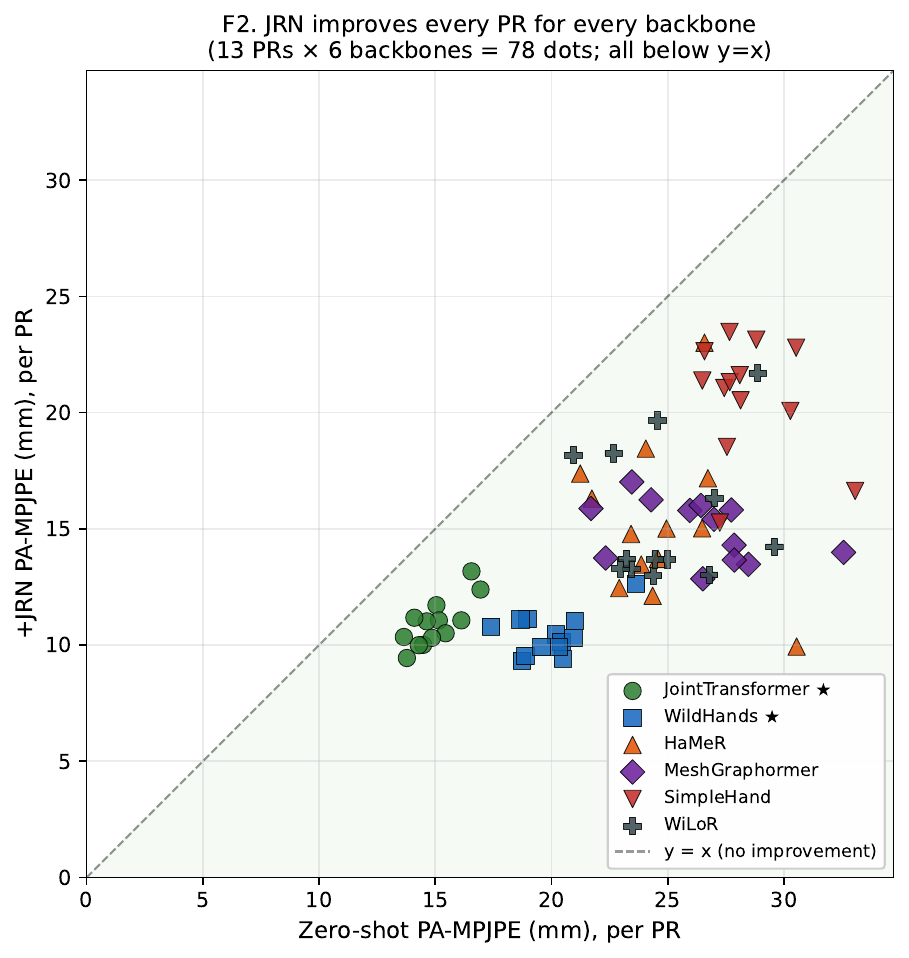}
  \caption{\textbf{Per-PR ZS$\to$+JRN scatter (78 dots: 13 PRs $\times$ 6 backbones).} Every dot lies below the y=x line: JRN improves accuracy on every PR for every backbone, with no negative regressions. Ego-pretrained backbones (green, blue) cluster at the lower-left.}
  \label{fig:scatter}
\end{figure}

\begin{table}[h]
  \centering
  \footnotesize
  \caption{\textbf{Per-stratum PA-MPJPE-MPS (mm) for ego-pretrained backbones with JRN.} P1-P4 = wearers; experience tiers MS1, PGY0, PGY3; glove colors blue, white, brown. All brown-glove frames are from P4; that confound is reported transparently here. The row-level $n$ reflects the wearer count; glove and experience cells aggregate across all wearers in the corresponding bucket (blue: P1-only, $n{=}4694$; white: P2+P3, $n{=}4121$; brown: P4-only, $n{=}661$; PGY3: P1+P4, $n{=}5355$; MS1: P2-only, $n{=}3198$; PGY0: P3-only, $n{=}923$). ``-'' marks cells anchored to a row above to avoid duplicating shared strata.}
  \label{tab:strata}
  \begin{tabular}{lrrrrrr}
    \toprule
     & \multicolumn{2}{c}{Wearer} & \multicolumn{2}{c}{Glove} & \multicolumn{2}{c}{Experience} \\
    \cmidrule(lr){2-3}\cmidrule(lr){4-5}\cmidrule(lr){6-7}
    Stratum (n) & WH+JRN & JT+JRN & WH+JRN & JT+JRN & WH+JRN & JT+JRN \\
    \midrule
    P1 / blue / PGY3 ($n{=}4694$) & 10.19 & 10.71 & 10.19 & 10.71 & 10.49 & 10.91 \\
    P2 / white / MS1 ($n{=}4121$/$3198$) & 10.19 & 10.26 & 10.28 & 10.42 & 10.19 & 10.26 \\
    P3 / - / PGY0 ($n{=}923$)    & 10.60 & 10.97 & -    & -    & 10.60 & 10.97 \\
    P4 / brown / PGY3 ($n{=}661$) & \textbf{12.64} & \textbf{12.39} & \textbf{12.64} & \textbf{12.39} & -    & -    \\
    \bottomrule
  \end{tabular}
\end{table}

\begin{figure}[h]
  \centering
  \includegraphics[width=\linewidth]{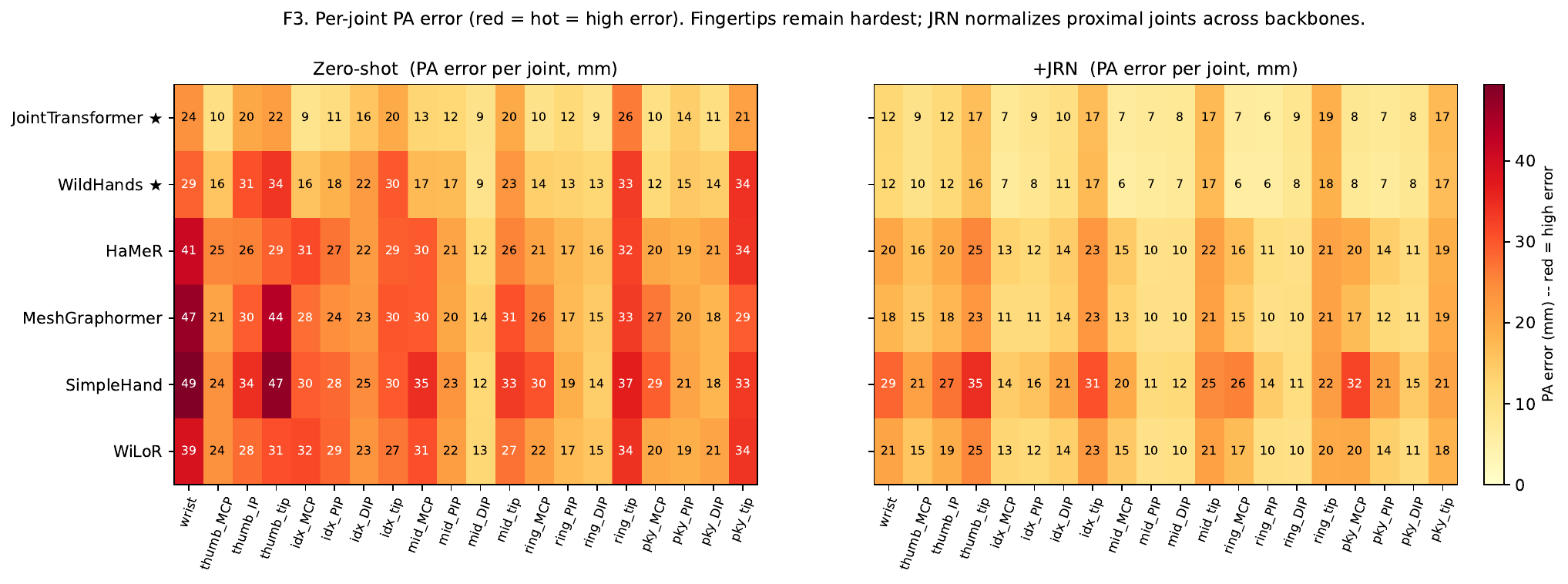}
  \caption{\textbf{Per-joint PA error heatmap (mm) for all 6 backbones, ZS vs +JRN.} Fingertips remain the hardest joints across all configurations. JRN normalizes proximal MCP joints across backbones (6-8\,mm post-correction for ego backbones).}
  \label{fig:per_joint_heatmap}
\end{figure}

\begin{figure}[h]
  \centering
  \includegraphics[width=\linewidth]{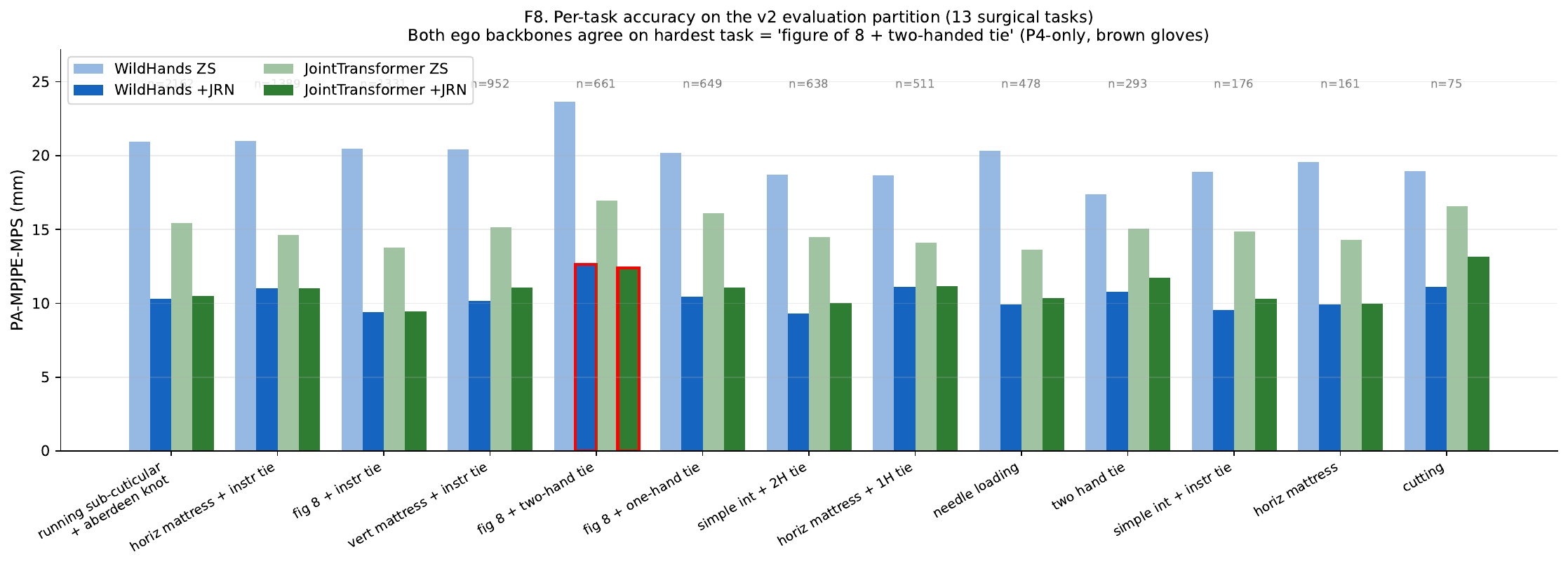}
  \caption{\textbf{Per-task PA-MPJPE-MPS on the evaluation partition.} 13 surgical training tasks; bars in pairs (ZS, +JRN) for WildHands ($\star$, blue) and JointTransformer ($\star$, green). Per-task $n$ shown above bars. Highlighted in red: \emph{figure of 8 + two-handed tie} (P4-only, brown-glove sessions; the hardest task for both ego backbones among tasks with $n{\geq}100$).}
  \label{fig:per_task}
\end{figure}

\begin{table}[h]
  \centering
  \footnotesize
  \caption{\textbf{Zero-shot and JRN-corrected PA-MPJPE for representative backbones.} All numbers PA-MPJPE-MPS in mm on the evaluation partition (9{,}476 hand instances). Zero-shot differences are the comparison surface; +JRN values sit at the label resolution scale (\S\ref{sec:labelquality}), so differences among them are not resolvable.}
  \label{tab:ablation}
  \begin{tabular}{lrl}
    \toprule
    Configuration & PA-MPJPE & Note \\
    \midrule
    HaMeR-ZS & 24.99 & non-egocentric, zero-shot \\
    WildHands-ZS & 20.48 & ego-pretrained, zero-shot \\
    HaMeR-ZS + JRN & 16.11 & within-corpus JRN \\
    WildHands-ZS + JRN & 10.40 & within-corpus JRN \\
    JointTransformer-ZS + JRN & 10.70 & within-corpus JRN \\
    MeshGraphormer-ZS + JRN & 15.12 & within-corpus JRN \\
    \bottomrule
  \end{tabular}
\end{table}

\begin{figure}[h]
  \centering
  \includegraphics[width=0.78\linewidth]{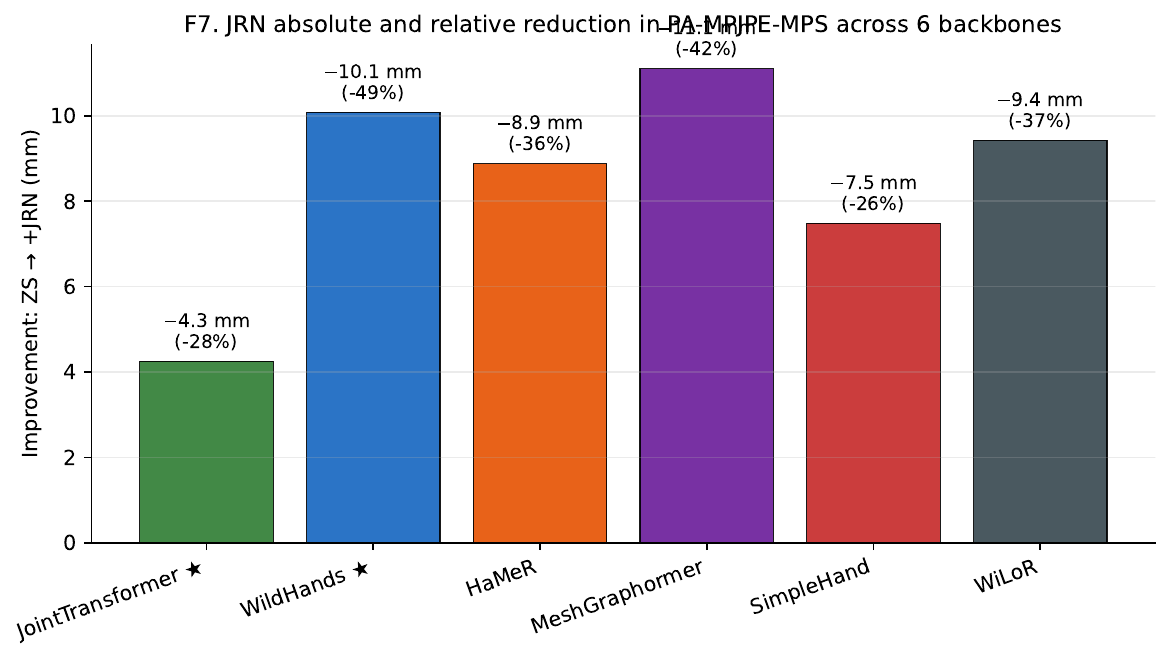}
  \caption{\textbf{JRN absolute and relative reduction in PA-MPJPE-MPS} for the 6 backbones. The WiLoR ZS$\to$JRN $\Delta$ shown reflects the post-fix WiLoR wrapper (Table~\ref{tab:headline}); a pre-fix joint-scramble artifact has been corrected.}
  \label{fig:delta}
\end{figure}

\begin{figure*}[t]
  \centering
  \includegraphics[width=\linewidth]{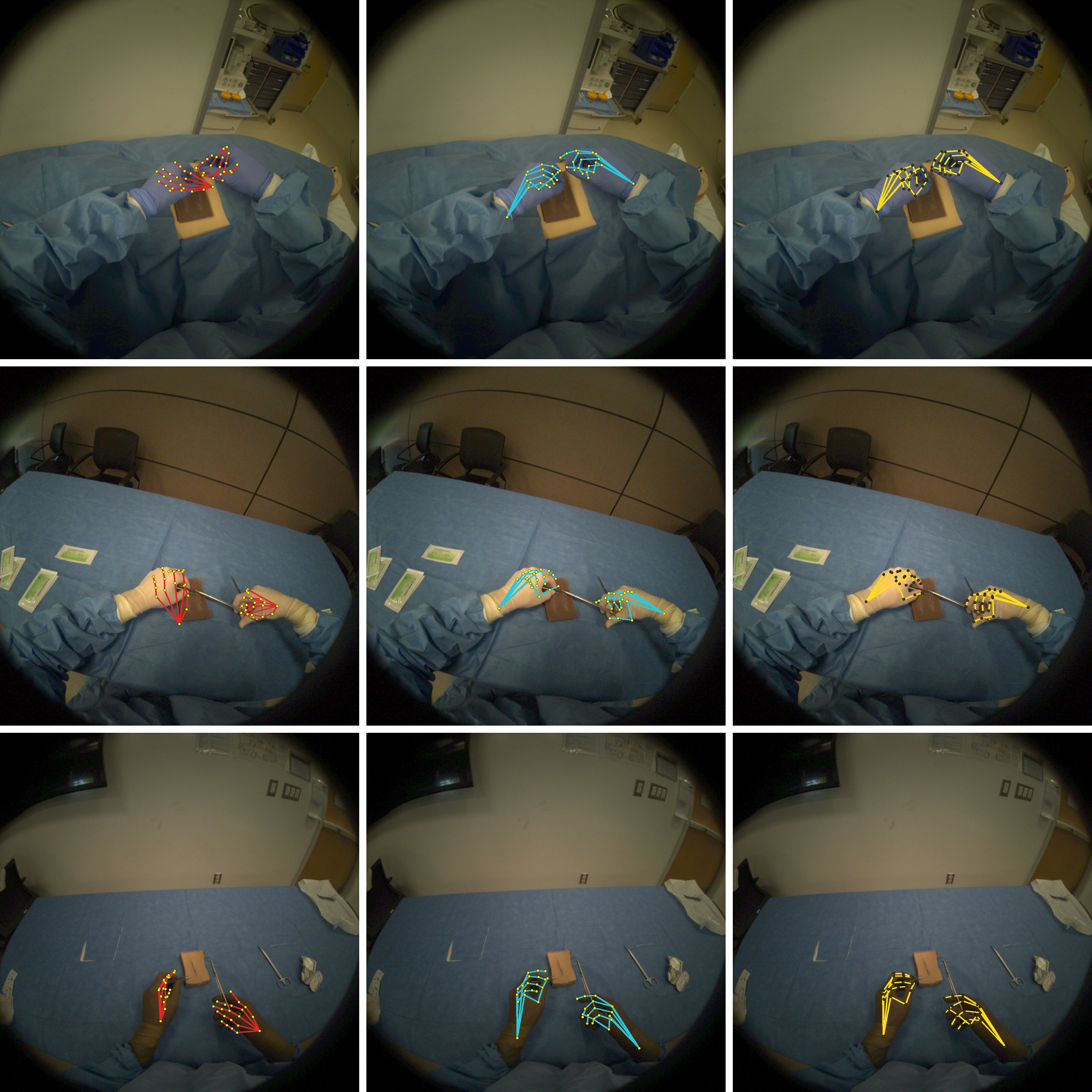}
  \caption{\textbf{Qualitative: WH-ZS vs WH+JRN vs MPS GT.} Three frames covering all glove conditions (top P1 blue, middle P2 white, bottom P4 brown). Columns: WildHands zero-shot, our WH+JRN-58K, MPS pseudo-GT. All overlays Procrustes-aligned to GT to isolate shape error. WH-ZS predicts an open-hand template that misses the surgical curl grip; JRN recovers it.}
  \label{fig:qual-zs-jrn-gt}
\end{figure*}

\clearpage
\section{External triangulated validation: details}
\label{app:extval}

This appendix details the validation summarized in \S\ref{sec:extval}.

\paragraph{Capture.} On one recording day, four wearers performed the training tasks in all three glove colors while two GoPro cameras (left and right obliques, 3840$\times$2160 at 59.94\,fps) and an overhead iPhone (3840$\times$2160 at 30\,fps; an iPhone 16 Pro for three wearers and an iPhone 12 Pro for the fourth) recorded alongside the Aria glasses, all started from one network control server. Of the 22 takes labeled by wearer and glove, 17 have all three external views and Aria SLAM.

\paragraph{Synchronization.} GoPro start times are read from the MP4 creation time, since the embedded timecode is offset by about 46\,s. The iPhone streams carry server-synchronized per-frame timestamps; each iPhone recording is aligned to each GoPro by cross-correlating per-frame image-motion energy, and the resulting left and right GoPro offsets agree within one 30\,fps frame. The GoPros are aligned to the Aria device clock by cross-correlating their audio with the Aria microphone. Audio and image motion give independent estimates of the left-to-right GoPro offset, which agree within 0 to 35\,ms on 16 of the 17 recordings; the exception is flagged below. MPS landmarks are interpolated at each external frame's capture time on the Aria clock.

\paragraph{Calibration.} A ChArUco board (10$\times$7 squares of 25\,mm, 18\,mm markers, DICT\_4X4\_50) lay on the table during the setup takes recorded the same morning and was removed before the labeled takes. The board poses show that the GoPros held position after setup (the left-to-right baseline direction agrees with every labeled recording within about 1 to 2$^\circ$), so we calibrate the two GoPros once as a shared rig. Left and right focal length and radial distortion, the left-to-right pose, and the board pose are estimated jointly with one iPhone pose per recording, by alternating triangulation and camera resection over the board corners (which fix the metric scale) and the hand keypoints of all recordings; the iPhone intrinsics come from the device. The rig has a 0.611\,m baseline, and board corners reproject at a median 3.3\,px. We chose this board-anchored rig in advance and report it for all recordings; per-recording self-calibration serves as a sensitivity check (Tab.~\ref{tab:extval-variants}). 

\paragraph{Keypoints and triangulation.} Hands are detected in each external view with the in-domain glove detector of our annotation pipeline, and the 2D keypoints are the average of HaMeR projections and RTMPose~\cite{jiang2023rtmpose} (hand) predictions on the same boxes. Hands are matched across views by triangulation consistency, preferring three-view matches, and triangulated from all matched views; most hands are seen by two views, as each oblique GoPro mainly sees the nearer hand. Triangulated hands are assigned to MPS hands by chirality (a left hand cannot be rigidly aligned to a right one) and then by proximity.

\paragraph{Frame selection and alignment.} From 79 instants per recording at a fixed interval, we keep hands with at least 15 matched joints, a median reprojection error of at most 25\,px, and MPS confidence of at least 0.97, then draw 500 frames by these quality criteria, balanced across the 11 wearer-glove strata the 17 recordings cover (13 to 51 frames per stratum). No step uses agreement with MPS. The reference is registered to the MPS world frame (MPS SLAM) by one robustly fitted similarity transform per recording (7 parameters); the rigid variant fixes the scale to the board's metric scale (6 parameters). Confidence intervals bootstrap over recordings.

\paragraph{Results.} Over 500 frames (562 hands, 9{,}552 joints), the median per-joint deviation is 15.9\,mm (95\% CI 14.7 to 17.5; IQR 10.7 to 24.3), of which the depth component along the Aria viewing ray is 9.1\,mm. With the rigid alignment it is 16.7\,mm (15.2 to 18.6), and MPS hands come out about 6\% smaller than the metric reference (median fitted scale 0.94). Deviation is lowest at the PIP and MCP joints (12 to 17\,mm) and highest at the fingertips (17 to 22\,mm) and the wrist (25\,mm), where the MPS and MANO joint definitions differ most. It varies little across wearers (14.4 to 18.5\,mm) and glove colors (15.4 to 16.4\,mm). After per-hand Procrustes alignment it is 9.8\,mm (per-recording medians 7.5 to 11.5\,mm, with one recording at 19.9\,mm). Tab.~\ref{tab:extval-variants} shows the result is stable across ways of building the reference.

\begin{table}[h]
\centering
\footnotesize
\caption{\textbf{External validation: per-joint deviation between the triangulated reference and MPS under different reference constructions} (500 frames each; similarity alignment unless stated).}
\label{tab:extval-variants}
\begin{tabular}{lr}
\toprule
Reference built with & Per-joint median, mm (95\% CI) \\
\midrule
Board-anchored shared rig, fused keypoints (reported) & 15.9 (14.7 to 17.5) \\
\quad same, rigid alignment with the board's metric scale & 16.7 (15.2 to 18.6) \\
Per-recording self-calibration, fused keypoints & 17.5 (16.4 to 18.7) \\
Per-recording self-calibration, HaMeR keypoints & 17.2 (16.0 to 18.5) \\
Per-recording self-calibration, RTMPose keypoints & 16.9 (15.4 to 18.2) \\
\bottomrule
\end{tabular}
\end{table}


\newpage
\input{checklist.tex}

\end{document}

%% file: checklist.tex
\section*{NeurIPS Paper Checklist}

\begin{enumerate}

\item {\bf Claims}
    \item[] Question: Do the main claims made in the abstract and introduction accurately reflect the paper's contributions and scope?
    \item[] Answer: \answerYes{}
    \item[] Justification: The abstract and introduction state the benchmark's scope (real-captured egocentric hand pose estimation in open-surgery training on synthetic pads; four wearers), the label type (sensor-fused, human-validated pseudo-ground truth, not mocap-grade), the zero-shot results with PR-keyed bootstrap CIs, the leave-one-wearer-out results, and JRN as a within-dataset diagnostic baseline. Each claim is stated with its scope beside it and is supported by Section~\ref{sec:experiments} (Tables~\ref{tab:headline} and~\ref{tab:lowo}, Sections~\ref{sec:iaa} and~\ref{sec:labelquality}).

\item {\bf Limitations}
    \item[] Question: Does the paper discuss the limitations of the work performed by the authors?
    \item[] Answer: \answerYes{}
    \item[] Justification: The Limitations paragraph (Section~\ref{sec:conclusion}) and the qualifications placed beside each claim disclose: (i) four wearers at a single site (ii) synthetic-skin pads rather than live-OR data, (iii) sensor-fused, human-validated pseudo-ground truth rather than mocap-grade labels, with a roughly 1\,cm label resolution scale below which corrected models cannot be separated (Section~\ref{sec:labelquality}), (iv) a within-corpus original split, with the leave-one-wearer-out study limited to five backbones and four wearers, (v) JRN as within-dataset MPS-label adaptation, and (vi) the wearer-confounded P4/brown stratum (Section~\ref{sec:lowo}).

\item {\bf Theory assumptions and proofs}
    \item[] Question: For each theoretical result, does the paper provide the full set of assumptions and a complete (and correct) proof?
    \item[] Answer: \answerNA{}
    \item[] Justification: The paper does not include formal theoretical results. The pose/translation orthogonality property of the JRN heads under Procrustes alignment is empirical (bit-identical PA-MPJPE across all 6 backbones) and a short derivation appears in Appendix~\ref{appx:orthogonal} as a sanity check rather than a theorem.

    \item {\bf Experimental result reproducibility}
    \item[] Question: Does the paper fully disclose all the information needed to reproduce the main experimental results of the paper to the extent that it affects the main claims and/or conclusions of the paper (regardless of whether the code and data are provided or not)?
    \item[] Answer: \answerYes{}
    \item[] Justification: Section~\ref{sec:experiments} and the JRN architecture appendix fully specify the model (58K-parameter two-headed MLP, zero-initialized heads), the optimizer (AdamW, lr$=10^{-3}$, weight decay $10^{-4}$, batch 512, 60 epochs, seed 42), the data partition (PR-disjoint, 41{,}909 train / 9{,}476 val; 42{,}956 is the pre-filter training partition, of which 1{,}047 rows are removed by the \texttt{has\_eval\_gt} valid-projection filter, leaving the released and benchmarked 41{,}909), the alignment metric (Umeyama Procrustes on the OP\_VALID 20-joint subset), and the bootstrap protocol (1{,}000 PR-keyed resamples). The released repository contains a single command that reproduces every headline number on a fresh A100 in under 10 seconds per backbone.

\item {\bf Open access to data and code}
    \item[] Question: Does the paper provide open access to the data and code, with sufficient instructions to faithfully reproduce the main experimental results, as described in supplemental material?
    \item[] Answer: \answerYes{}
    \item[] Justification: The dataset is released on HuggingFace as parquet shards with an accompanying README, datasheet (Datasheets for Datasets template), and Croissant JSON-LD metadata (NeurIPS 2026 E\&D requirement). The code repository (\url{https://github.com/monk1337/egoSurgHands}) provides JRN training/eval scripts, the annotation tool, and a single-command reproduction path. URLs are listed in the abstract.

\item {\bf Experimental setting/details}
    \item[] Question: Does the paper specify all the training and test details (e.g., data splits, hyperparameters, how they were chosen, type of optimizer) necessary to understand the results?
    \item[] Answer: \answerYes{}
    \item[] Justification: Splits (Section~\ref{sec:splits}), evaluation metrics with explicit alignment and joint-subset definitions (Section~\ref{sec:metrics}), and the experimental protocol (Section~\ref{sec:protocol}) are stated in the main paper. The JRN architecture appendix lists all hyperparameters; no hyperparameter search was performed (single seed, single run, no early stopping).

\item {\bf Experiment statistical significance}
    \item[] Question: Does the paper report error bars suitably and correctly defined or other appropriate information about the statistical significance of the experiments?
    \item[] Answer: \answerYes{}
    \item[] Justification: All headline metrics in Table~\ref{tab:headline} and the forest plot (Figure~\ref{fig:forest_combined}) report 95\% bootstrap confidence intervals from 1{,}000 PR-keyed resamples (PR is the natural unit of correlation, $n=13$ PRs in the val partition). The bootstrap method, resampling unit, and seed are stated in Section~\ref{sec:metrics}. CIs are non-symmetric where appropriate, and they describe uncertainty over the 13 available evaluation PRs, not evidence of broader generalization.

\item {\bf Experiments compute resources}
    \item[] Question: For each experiment, does the paper provide sufficient information on the computer resources (type of compute workers, memory, time of execution) needed to reproduce the experiments?
    \item[] Answer: \answerYes{}
    \item[] Justification: Each JRN training run is approximately 8 seconds on a single A100 80GB GPU (training data fits in GPU memory: $41{,}909 \times 66 \times 4$\,B $\approx 11$\,MB); the full headline matrix (6 backbones $\times$ JRN) completes in under 2 minutes. Backbone zero-shot inference uses public checkpoints. Total project compute (including the parquet build pipeline, baseline ZS evaluation, and ablations) is reported in the appendix.

\item {\bf Code of ethics}
    \item[] Question: Does the research conducted in the paper conform, in every respect, with the NeurIPS Code of Ethics \url{https://neurips.cc/public/EthicsGuidelines}?
    \item[] Answer: \answerYes{}
    \item[] Justification: All recordings were captured under informed consent of the four wearers, in a non-clinical simulation lab using synthetic-skin training pads (no patients), with wearer identities anonymized to P1--P4 throughout the dataset and paper. The release license is non-commercial, and the data includes no personally identifiable information (50/50 sample audit confirmed; see Appx.~\ref{app:rai}).

\item {\bf Broader impacts}
    \item[] Question: Does the paper discuss both potential positive societal impacts and negative societal impacts of the work performed?
    \item[] Answer: \answerYes{}
    \item[] Justification: Appx.~\ref{app:rai} and the Conclusion (Section~\ref{sec:conclusion}) discuss positive impacts (objective surgical-skill assessment, AR-guided training feedback) and negative impacts (premature clinical deployment, distribution shift to live-OR, mis-application to high-stakes decisions). The paper explicitly states that trained models are not for clinical decision-making without further validation.

\item {\bf Safeguards}
    \item[] Question: Does the paper describe safeguards that have been put in place for responsible release of data or models that have a high risk for misuse (e.g., pre-trained language models, image generators, or scraped datasets)?
    \item[] Answer: \answerYes{}
    \item[] Justification: The dataset is released under CC BY-NC-SA 4.0 (non-commercial, share-alike), restricting downstream commercial deployment. No facially identifiable persons appear in the recorded RGB (verified via 50-frame manual audit and the camera geometry pointing down at the surgical pad), and wearer IDs are anonymized. The accompanying datasheet and Croissant metadata document intended use, restrictions, and known limitations.

\item {\bf Licenses for existing assets}
    \item[] Question: Are the creators or original owners of assets (e.g., code, data, models), used in the paper, properly credited and are the license and terms of use explicitly mentioned and properly respected?
    \item[] Answer: \answerYes{}
    \item[] Justification: All baselines (HaMeR, WildHands, JointTransformer, MeshGraphormer, SimpleHand, WiLoR, IntagHand, MediaPipe Hands) are cited in Section~\ref{sec:baselines} and used under their respective public licenses. Aria Project hardware and MPS outputs are credited and used under the Aria Research Kit license. The MANO model is cited and used under its non-commercial research license. \texttt{projectaria\_tools} is Apache-2.0. Per-asset licenses are listed in the supplementary.

\item {\bf New assets}
    \item[] Question: Are new assets introduced in the paper well documented and is the documentation provided alongside the assets?
    \item[] Answer: \answerYes{}
    \item[] Justification: Three new assets are released alongside the paper: (i) the EgoSurgHands dataset with parquet shards, README, datasheet (Gebru et al., 2021), and Croissant metadata; (ii) the JRN model definition and training script; (iii) the open-source IAA annotation tool. All include license, intended use, and limitations documentation. URLs are listed in the abstract.

\item {\bf Crowdsourcing and research with human subjects}
    \item[] Question: For crowdsourcing experiments and research with human subjects, does the paper include the full text of instructions given to participants and screenshots, if applicable, as well as details about compensation (if any)?
    \item[] Answer: \answerYes{}
    \item[] Justification: The Inter-Annotator-Agreement (IAA) workflow described in Section~\ref{sec:iaa} involves lab-internal reviewers (primary review, examiner review, and senior-surgeon adjudication, with three raters on the 500-frame agreement audit; not external crowdworkers) verifying MPS-projected hand-pose overlays via the open-source annotation tool. Reviewer instructions, the keyboard-shortcut reference, and screenshots of the annotation interface are included in the supplementary. Reviewers were lab members; no external paid crowdsourcing was used.

\item {\bf Institutional review board (IRB) approvals or equivalent for research with human subjects}
    \item[] Question: Does the paper describe potential risks incurred by study participants, whether such risks were disclosed to the subjects, and whether Institutional Review Board (IRB) approvals (or an equivalent approval/review based on the requirements of your country or institution) were obtained?
    \item[] Answer: \answerYes{}
    \item[] Justification: Recordings used synthetic-skin surgical-training pads with consenting volunteer wearers (no patients, no patient data); formal IRB review was not required under the institution's non-human-subjects-research policy (RAI appendix).

\item {\bf Declaration of LLM usage}
    \item[] Question: Does the paper describe the usage of LLMs if it is an important, original, or non-standard component of the core methods in this research? Note that if the LLM is used only for writing, editing, or formatting purposes and does \emph{not} impact the core methodology, scientific rigor, or originality of the research, declaration is not required.
    \item[] Answer: \answerNA{}
    \item[] Justification: Large language models were used only for editing and formatting assistance. They are not part of the core methodology: the dataset capture pipeline, the JRN architecture, the training procedure, and the empirical results are produced by deterministic non-LLM components.

\end{enumerate}